\documentclass[letterpaper]{article} 
\usepackage{aaai23}  
\usepackage{times}  
\usepackage{helvet}  
\usepackage{courier}  
\usepackage[hyphens]{url}  
\usepackage{graphicx} 
\def\UrlFont{\rm}  
\usepackage{natbib}  
\usepackage{caption} 
\usepackage{algorithm}
\usepackage{algorithmic}

\usepackage{newfloat}
\usepackage{listings}
\DeclareCaptionStyle{ruled}{labelfont=normalfont,labelsep=colon,strut=off} 
\floatstyle{ruled}
\newfloat{listing}{tb}{lst}{}
\floatname{listing}{Listing}
\usepackage{tikz}
\usepackage{comment}
\usepackage{amsmath,amssymb} 
\usepackage{multirow}
\usepackage{makecell}
\usepackage{bbding}
\usepackage{subcaption}
\usepackage{xspace}
\usepackage{kotex}

\newcommand{\eg}{\textit{e.g.}\xspace}
\newcommand{\ie}{\textit{i.e.}\xspace}
\newcommand{\etal}{\textit{et al.}\xspace}

\usepackage{color}
\usepackage{xcolor}
\definecolor{cadmiumred}{rgb}{0.89, 0.0, 0.13}
\newcommand{\red}[1]{#1}

\makeatletter

\newcommand*{\@rowstyle}{}

\newcommand*{\rowstyle}[1]{
  \gdef\@rowstyle{#1}%
  \@rowstyle\ignorespaces%
}

\newcolumntype{=}{
  >{\gdef\@rowstyle{}}%
}

\newcolumntype{+}{
  >{\@rowstyle}%
}

\usepackage{xr} 
\makeatletter
\newcommand*{\addFileDependency}[1]{
  \typeout{(#1)}
  \@addtofilelist{#1}
  \IfFileExists{#1}{}{\typeout{No file #1.}}
}
\makeatother

\usepackage{algorithm}
\usepackage{listings}
\title{Frequency Selective Augmentation for Video Representation Learning}

\author {
    Jinhyung Kim\textsuperscript{\rm 1}\thanks{This work is done during Ph.D. program at KAIST.},
    Taeoh Kim\textsuperscript{\rm 2},
    Minho Shim\textsuperscript{\rm 2},
    Dongyoon Han\textsuperscript{\rm 3},
    Dongyoon Wee\textsuperscript{\rm 2},
    Junmo Kim\textsuperscript{\rm 4}
}
\affiliations {
    \textsuperscript{\rm 1} LG AI Research,
    \textsuperscript{\rm 2} NAVER CLOVA Video,
    \textsuperscript{\rm 3} NAVER AI Lab,
    \textsuperscript{\rm 4} KAIST\\
}

\begin{document}

\maketitle

\begin{abstract}
Recent self-supervised video representation learning methods focus on maximizing the similarity between multiple augmented views from the same video and largely rely on the quality of generated views. 
However, most existing methods lack a mechanism to prevent representation learning from bias towards static information in the video.
In this paper, we propose frequency augmentation (FreqAug), a spatio-temporal data augmentation method in the frequency domain for video representation learning.
FreqAug stochastically removes specific frequency components from the video so that learned representation captures essential features more from the remaining information for various downstream tasks. 
Specifically, FreqAug pushes the model to focus more on dynamic features rather than static features in the video via dropping spatial or temporal low-frequency components.
To verify the generality of the proposed method, we experiment with FreqAug on multiple self-supervised learning frameworks along with standard augmentations.
Transferring the improved representation to five video action recognition and two temporal action localization downstream tasks shows consistent improvements over baselines.
\end{abstract}

\section{Introduction}
\label{sec:intro}
There has been growing attention on transferring knowledge from large-scale unsupervised learning to various downstream tasks in natural language processing~\cite{devlin2018bert,radford2019language} and computer vision~\cite{Chen2020simclr,He2020moco,Jean2020byol} communities.
Considering data accessibility and possible applications, video representation learning has great potential as a tremendous amount of videos with diverse contents are created, shared, and consumed every day.
In fact, unsupervised or self-supervised learning (SSL) of video via learning invariance between multimodal or multiple augmented views of an instance is being actively studied \cite{Han20coclr,Alayrac20mmv,recasen21brave,huang2021self,Qian2021CVRL,Feichtenhofer2021largescale}.

Recent studies in image SSL indicate that a careful selection of data augmentation is crucial for the quality of the feature~\cite{wen2021sparsefeature} or for improving performance in downstream tasks~\cite{Tian2020goodview,Zhao2021goodtransfer}.
However, augmentations for video SSL have not been sufficiently explored yet.
For videos, in terms of spatial dimension, the standard practice is adopting typical image augmentations in a temporally consistent way, \ie, applying the same augmentation to every frame~\cite{Qian2021CVRL}.
Meanwhile, a few previous works have investigated augmentations in the temporal dimension, including sampling a distant clip~\cite{Feichtenhofer2021largescale}, sampling clips with different temporal scales~\cite{dave2021tclr} or playback speeds~\cite{chen2020RSPNet,Huang2021ascnet,duan2022transrank}, and dropping certain frames~\cite{pan2021videomoco}.
Although effective, sampling-based augmentations in the temporal dimension inevitably modulate a video as a whole regardless of signals in a clip varying at different rates.
These methods are limited in resolving the spatial bias problem~\cite{Li2018RESOUND} of video datasets which requires distinguishing motion-related features from static objects or scenes.
Adding a static frame~\cite{wang2021removing} is a simple heuristic to attenuate the temporally stationary signal, but it is hard to generalize to the real world's non-stationary signal in the spatio-temporal dimension.
The need for a more general way to selectively process a video signal depending on the spatial and temporal changing rates motivates us to consider frequency domain analysis.

In digital signal processing, converting a signal to the frequency domain using discrete Fourier transform (DFT), then processing the signal is widely used in many applications.
Filtering in the frequency domain is one example that attenuates a specific frequency range to remove undesirable components, such as noise, from the signal.
With its effectiveness in mind, we propose filtering video signals in the frequency domain to discard unnecessary information while keeping desired features for the SSL model to learn.

\begin{figure}[t]
  \centering
   \includegraphics[width=0.9\linewidth]{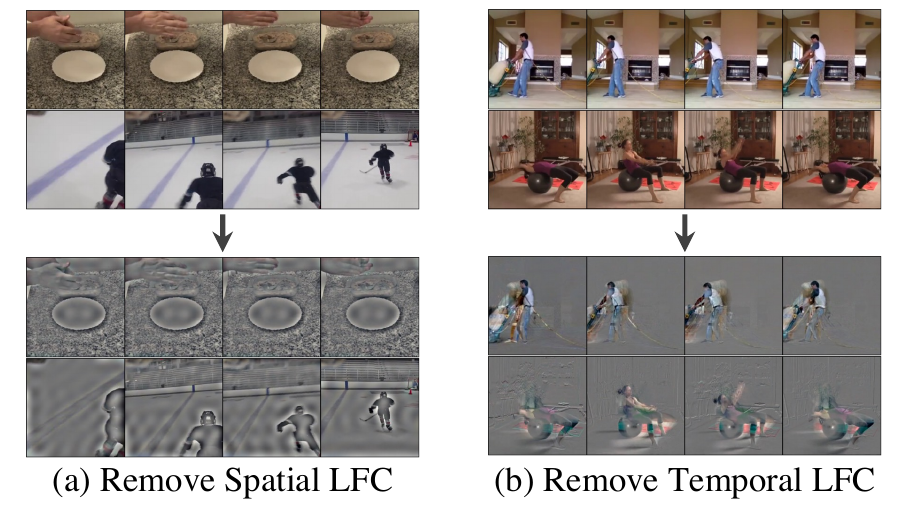}
   \caption{\textbf{Impact of removing low-frequency components (LFC).} (a) Filtering spatial LFC can attenuate spatially redundant information, \eg colors, while keeping the shape patterns. (b) Removing temporal LFC filters out temporally stationary information, \eg the background, while keeping the motion pattern. 
   }
   \label{fig:teaser}
\end{figure}
Fig.~\ref{fig:teaser} shows the outcome of filtering out low-frequency components (LFC) from videos.
When the spatial filter is applied, plain surfaces of objects in the scene are erased while their boundary or shapes are remained.
As for the temporal filter, stationary parts of the video, \eg, static objects or the background, are removed while dynamic parts, \eg, a person moving, are retained.
These results are aligned with the previous discoveries that high-frequency components (HFC) carry essential information for image and video understanding~\cite{Wang2020highfreq,Kim2020rms}.

In this work, we propose frequency augmentation (FreqAug), a novel spatio-temporal augmentation in the frequency domain by randomly applying a filter to remove selective frequency bands from the video.
Specifically, we aim to alleviate representation bias for better transferability by filtering out spatially and temporally static components from the video signal.
FreqAug is composed of a 2D spatial filter and a 1D temporal filter, and their frequency band can be determined by the filter type and its cutoff frequency.
In video SSL, FreqAug can be applied to each view independently on top of other video augmentations so that the model learns invariance on LFC (Fig.~\ref{fig:tsne}).
In particular, applying FreqAug with high-pass filter results in obtaining the representation with less static bias via learning invariant features between the instance and its HFC.
Note that what we are claiming is not that only HFC are important but rather a matter of relative importance.
Since FreqAug is applied randomly, LFC still get involved in the invariance learning. 

We demonstrate the effectiveness of the proposed method by presenting transfer learning performance on five action recognition datasets: coarse-grained (UCF101 and HMDB51) and fine-grained (Diving48, Gym99, and Something-Something-v2) datasets.
Additionally, the learned features are evaluated via the temporal action segmentation task on Breakfast dataset and the action localization task on THUMOS'14 dataset.
Empirical results show that FreqAug enhances the performance of multiple SSL frameworks and backbones, which implies the learned representation has significantly improved transferability. 
\red{We also make both quantitative and qualitative analyses of how FreqAug can affect video representation learning. }

\section{Related Work}
\label{sec:related}
\subsection{Frequency Domain Augmentations}
\label{subsec:ft}
Lately, several studies on frequency domain augmentation have been proposed for the 1D speech and 2D image domains.
For speech or acoustic signals, a few works incorporated augmentations that are masking~\cite{park2019specaugment} or filtering~\cite{nam2021filteraugment} a spectrogram or mixing that of two samples~\cite{kim2021specmix}.   
In the image domain, Xu \etal ~\cite{Xu2021FACT} tackled the domain generalization problem by mixing spectrum amplitude of two images. 
A concurrent work~\cite{Nam2021FF} introduced randomly masking a certain angle of Fourier spectrum based on the spectrum intensity distribution for X-ray image classification. 
These methods are relevant to ours in that they randomly alter a spectrum for data augmentation, but differs in the following two respects. 
\red{
First, to the best of our knowledge, our work is the first 3D spatio-temporal augmentation in the frequency domain for video representation learning and investigates the transferability to various downstream tasks. 
Second, our method differs from the existing frequency domain augmentations in that it selectively filters out a certain frequency band, \eg, low-frequency components, rather than random frequency components through the entire range.
We empirically show the superiority of selective filtering over the random filtering strategy (Table~\ref{table:filter}).
}

\subsection{Video Self-Supervised Learning}
\label{subsec:videossl}
Self-supervised learning (SSL) through multi-view invariance learning has been widely studied for image recognition and other downstream tasks \cite{Wu2018unsupervised,Chen2020simclr,He2020moco,Chen2020mocov2,Jean2020byol,Caron2020swav,Chen2021simsiam}.
In video SSL, previous works exploited the view invariance-based approaches from the image domain and explored ways to utilize unique characteristics of the video including additional modalities, \eg, optical flow, audio, and text~\cite{wang2021enhancing,huang2021self,xiao2021modist,Han20coclr,miech19milnce,Alayrac20mmv,alwassel2020xdc,recasen21brave,Behrmann2021longshortview}.
However, we focus more on the RGB-based video SSL methods in this study. 
CVRL~\cite{Qian2021CVRL} proposed a temporally consistent spatial augmentation and temporal sampling strategy, which samples two positive clips more likely from near time.
RSPNet~\cite{chen2020RSPNet} combined relative speed perception and video instance discrimination tasks to learn both motion and appearance features from video. 
Empirical results in~\cite{Feichtenhofer2021largescale} show four image-based SSL frameworks~\cite{Chen2020mocov2,Jean2020byol,Chen2020simclr,Caron2020swav} can be generalized well to the video domain.
MoCo-BE~\cite{wang2021removing} and FAME~\cite{ding2022fame} introduced a regularization that reduces background influences on SSL by adding a static frame to the video or mixing background, respectively.
Suppressing static cues~\cite{zhang2022suppressing} in the latent space is another way to reduce spatial bias.
Our work is also a study on data augmentation for video SSL, but we propose to modulate the video signal in the frequency domain in a more general and simpler way.

\section{Method}
\label{sec:method}
\subsection{Preliminary}
\label{subsec:dft}
In this work, we aim to augment spatio-temporal video signals in a frequency domain by filtering particular frequency components.
Discrete Fourier transform (DFT), a widely used technique in many digital signal processing applications, provides appropriate means of converting a finite discrete signal into the frequency domain for computers.
For simplicity, let us consider 1D discrete signal $x[n]$ of length $N$, then 1D DFT is defined as,
\begin{equation}
  X[k]=\sum_{n=0}^{N-1} x[n]e^{-j(2 \pi /N)kn},
  \label{eq:dft}
\end{equation}
where $X[k]$ is the spectrum of $x[n]$ at frequency $k=0,1,...,N-1$.
Since DFT is a linear transformation, the original signal can be reconstructed by inverse discrete Fourier transform (iDFT):
\begin{equation}
  x[n]=\frac{1}{N}\sum_{k=0}^{N-1} X[k]e^{j(2 \pi /N)kn}.
  \label{eq:idft}
\end{equation}

1D-DFT can be extended to the multidimensional DFT by simply calculating a series of 1D-DFT along each dimension.
One can express d-dimensional DFT in a concise vector notation as,
\begin{equation}
  X_\mathbf{k}=\sum_{\mathbf{n}=0}^{\mathbf{N}-1} x_\mathbf{n}e^{-j2 \pi\mathbf{k}(\mathbf{n/N})},
  \label{eq:ddft}
\end{equation}
where $\mathbf{k}=(k_1,k_2,...,k_d)$ and $\mathbf{n}=(n_1,n_2,...,n_d)$ are d-dimensional indices from $\mathbf{0}$ to $\mathbf{N}=(N_1,N_2,...,N_d)$ and $\mathbf{n/N}$ is defined as $(n_1/N_1,n_2/N_2,...,n_d/N_d)$. We omit the equation of d-dimensional iDFT as it is a straightforward modification from Eq.~\ref{eq:idft}.


\subsection{Filtering Augmentation in Frequency Domain}
\label{subsec:filtering}
Filtering in signal processing often denotes a process of suppressing certain frequency bands of a signal.
Filtering in frequency domain can be described as an element-wise multiplication $\odot$ between a filter $F$ and a spectrum $X$ as,
\begin{equation}
  \hat{X} = F\odot X,
  \label{eq:filtering}
\end{equation}
where $\hat{X}$ is a filtered spectrum.
A filter can be classified based on the frequency band that the filter passes or rejects: low-pass filter (LPF), high-pass filter (HPF), band-pass filter, band-reject filter, and so on.
LPF passes a low-frequency band while it filters out high-frequency components from the signal; HPF works oppositely.
Let us consider a simple 1D binary filter, also known as an ideal filter, then LPF and HPF can be defined as,
\begin{equation}
  F_{lpf}[k] = 
    \begin{cases}
        1 & \text{if $|k|<f_{co}$}\\
        0 & \text{otherwise},
    \end{cases}
  \label{eq:filter1d_lpf}
\end{equation}
\begin{equation}
  F_{hpf}[k] = 1 - F_{lpf}[k],
  \label{eq:filter1d_hpf}
\end{equation}
where $f_{co}$ is the cutoff frequency which controls the frequency band of the filter. 


\begin{figure}[t]
\centering
    \includegraphics[width=0.95\linewidth]{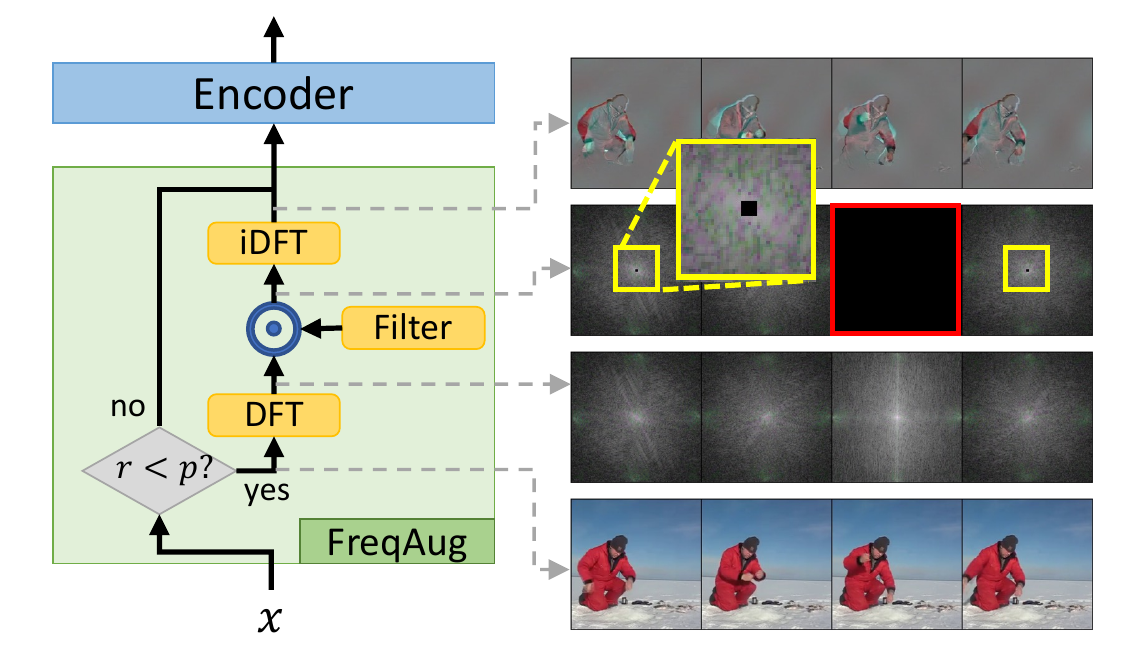}
    \caption{\textbf{Frequency augmentation (FreqAug).}  Filtering in the frequency domain is a sequential process of 1) transforming a video to a spectrum by DFT; 2) applying the desired filter by element-wise multiplication; 3) transforming the filtered spectrum back to the video domain by iDFT. Figures on the right are an example of applying spatio-temporal high-pass filters. In filtered spectrum (2nd row), low-frequency components of spatial (small black regions inside yellow boxes, see the first one for the close-up) and temporal (the red box) axis are removed. FreqAug is placed after other augmentations and randomly applied when $r{\sim}U(0,1)$ is less than the augmentation probability $p$.}
\label{fig:freqfilter}
\end{figure}
In this work, we propose frequency augmentation (FreqAug, Fig.~\ref{fig:freqfilter}), which utilizes 3D-DFT with the binary filter approach to augment video data in the frequency domain by stochastically removing certain frequency components.
Since video signals have three dimensions, \ie, T, H, and W, the filter also can be 3D and have three independent cutoff frequencies.
We introduce a single spatial cutoff frequency $f^s_{co}$ that handles both H and W dimension, and one temporal cutoff frequency $f^t_{co}$ for T dimension.
Then 1D temporal filters are identical to Eq.~\ref{eq:filter1d_lpf} and Eq.~\ref{eq:filter1d_hpf}, and 2D spatial LPF can be defined as,
\begin{equation}
  F^{s}_{lpf}[k_h, k_w] = 
    \begin{cases}
        1 & \text{if $|k_h|<f^s_{co}$ and $|k_w|<f^s_{co}$}\\
        0 & \text{otherwise},
    \end{cases}
  \label{eq:filter2d_lpf}
\end{equation}
and $F^{s}_{hpf}$ is obtained in the same way as Eq.~\ref{eq:filter1d_hpf}.
Finally, the spatio-temporal filter $\mathbf{F}$ can be obtained by outer product between the temporal filter $F^t$ and the spatial filter $F^s$ as, 
\begin{equation}
  \mathbf{F}=F^{st}[k_t, k_h, k_w]=F^t[k_t]\otimes F^s[k_h, k_w],
  \label{eq:filter3d}
\end{equation}
where $\otimes$ is outer product.
The final 3D filtered spectrum $\hat{\mathbf{X}}$ can be represented as an element-wise multiplication between $\mathbf{F}$ and the spectrum $\mathbf{X}$ as Eq.~\ref{eq:filtering}.

Additionally, FreqAug has one more hyperparameter, the augmentation probability $p$, which determines how frequently the augmentation is applied.
FreqAug processes the input only when the random scalar $r$, sampled from uniform distribution $U(0,1)$, is less than $p$.

Fig.~\ref{fig:freqfilter} presents a block diagram of FreqAug and a visualization of a video sample and its spectrum at each stage of FreqAug.
Note that FreqAug blocks are located after other augmentations or noramlization, and operate with independent $r$ for each view. 
For the spectrum, lower absolute spatial frequencies are located near the center of the spectrum at each column ($(k_h, k_w)=(0, 0)$) and lower absolute temporal frequencies are located near the third spectrum ($k_t=0$).
For visualization, we apply spatial and temporal HPF with $f^s_{co}=0.01$ and $f^t_{co}=0.1$, respectively.
In the filtered spectrum (2nd row), spatial low-frequency (small black region inside yellow boxes) and temporal low-frequency (red box) components are removed.

\section{Experiment}
\label{sec:exp}

\subsection{Experiment Settings} 
Here, we provide essential information to understand the following experiments. Refer to Appendix~\ref{sec:ax_implementation} for more details.

\noindent
\textbf{Datasets.}
\red{ 
For pretraining the model, we use Kinetics-400 (K400)~\cite{Carreira2017QuoVA} and Mini-Kinetics (MK200)~\cite{Xie2018rethinking}. 
With the limited resources, we choose MK200 as a major testbed to verify our method's effectiveness.
}
\red{ 
For evaluation of the pretrained models, we use five different action recognition datasets: UCF101~\cite{soomro2012ucf101}, HMDB51~\cite{kuehne2011hmdb}, Diving48~\cite{Li2018RESOUND}, Gym99~\cite{shao2020finegym}, and Something-something-v2 (SSv2)~\cite{Goyal2017something}.
Following the standard practice, we report the finetuning accuracy on the three datasets: UCF101, HMDB51, and Diving48.
Note that we present split-1 accuracy for UCF101 and HMDB51 by default unless otherwise specified.
For Gym99 and SSv2, we evaluate the models on the low-shot learning protocol using only 10\% of training data since they are relatively large-scale (especially the number of samples in SSv2 is about twice larger than that of our main testbed MK200).
For temporal action localization, Breakfast~\cite{Kuehne12breakfast} and THUMOS'14~\cite{idrees2017thumos} dataset are used.
}

\noindent
\textbf{Self-supervised Pretraining.}
For self-supervised pretraining, all the models are trained with SGD for 200 epochs.
Regarding spatial augmentation, augmentations described in~\cite{Chen2020mocov2} are applied as our baseline. 
For temporal augmentation, randomly sampled clips from different timestamps compose the positive instances.
Also, two clips are constrained to be sampled within a range of 1 second.
Each clip consists of $T$ frames sampled from $T\times \tau$ consecutive frames with the stride $\tau$.
In terms of FreqAug, we use the following two default settings: 1) FreqAug-T (temporal) uses temporal HPF with a cutoff frequency 0.1; 2) FreqAug-ST (spatio-temporal) is a combination of spatial HPF with a cutoff frequency 0.01 alongside with FreqAug-T.

\noindent
\textbf{Finetuning and Low-shot Learning.}
We train the models for 200 epochs with the initial learning rate 0.025 without warm-up and zeroed weight decay for supervised finetuning and low-shot learning.
Only fundamental spatial augmentations~\cite{Feichtenhofer2021largescale} are used.

\noindent
\textbf{Temporal Action Segmentation and Localization.}
We train an action segmentation model, MS-TCN~\cite{Farha2019mstcn} following~\cite{Behrmann2021longshortview}, and a localization model, G-TAD~\cite{xu2020g} for evaluating the learned representation of pretrained encoders.

\noindent
\textbf{Evaluation.}
For Kinetics, UCF101, and HMDB51, we report average accuracy over 30-crops following~\cite{Chris2019slowfast}.
In the case of Diving48, Gym99, and SSv2, we report the spatial 3-crop accuracy with segment-based temporal sampling.
For temporal action segmentation, frame-wise accuracy, edit distance, and F1 score at overlapping thresholds 10\%, 25\%, and 50\% are used.
For temporal action localization, we measure mean average precision (mAP) with intersection-over-union (IoU) from 0.3 to 0.7.

\noindent
\textbf{Backbone.}
Our default encoder backbone is SlowOnly-50 (SO-50), a variant of 3D ResNet originated from the slow branch of SlowFast Network~\cite{Chris2019slowfast}.
We evaluate our method on R(2+1)D~\cite{Tran2018r21d} and S3D-G~\cite{Xie2018rethinking} models as well.

\noindent
\textbf{SSL Methods.}
We implement MoCo~\cite{Chen2020mocov2} and BYOL~\cite{Jean2020byol} for 
 pretraining the video model. We set MoCo as our default SSL method.

\subsection{Action Recognition Evaluation Results}
\label{subsec:mk200}
\begin{table}[!t]
\tabcolsep=0.07cm
\begin{center}
\resizebox{0.95\linewidth}{!}{
		\begin{tabular}{c|c|l||c|c|c|c|c}
            \hline
            \multirow{2}{*}{\makecell{Backbone }} & \multirow{2}{*}{\makecell{$T\times \tau$ }} &
             \multirow{2}{*}{\makecell{Augment. }} & \multicolumn{3}{c|}{Finetune} &\multicolumn{2}{c}{Low-shot (10\%)}\\
            \cline{4-8}
            & &  & UCF101 & HMDB51 & Diving48 & Gym99 & SSv2 \\
            \hline
              \multirow{3}{*}{\makecell{SO-50}} & \multirow{3}{*}{\makecell{$8\times 8$}} &
              Baseline & 87.0 & 56.5 & 67.8 & 29.9 & 25.3 \\
             & &  + \textbf{FA-ST} & \textbf{90.0} & \textbf{61.6} & \textbf{71.0} & 34.8 & \textbf{28.1} \\
             & &  + \textbf{FA-T} & 89.8 & 60.8 & 70.3 & \textbf{35.2} & \textbf{28.1} \\
            \hline
              \multirow{3}{*}{\makecell{SO-18}} & \multirow{3}{*}{\makecell{$16\times 4$}} &
              Baseline & 84.5 & 55.2 & 74.9 & 30.3 & 23.9 \\
             & &  + \textbf{FA-ST} & 88.5 & 57.8 & \textbf{75.8} & \textbf{35.3} & 25.7 \\
             & &  + \textbf{FA-T} & \textbf{88.7} & \textbf{58.8} & 75.7 & 34.7 & \textbf{26.1} \\
            \hline
              \multirow{3}{*}{\makecell{R(2+1)D}} & \multirow{3}{*}{\makecell{$32\times 2$}} &
              Baseline & 86.2 & 60.4 & 64.6 & 42.5 & 29.2 \\
             & &  + \textbf{FA-ST} & \textbf{90.0} & \textbf{65.9} & 67.7 & \textbf{48.4} & \textbf{31.5} \\
             & &  + \textbf{FA-T} & 89.5 & 65.2 & \textbf{70.2} & 48.3 & 30.5 \\
            \hline
             \multirow{3}{*}{\makecell{S3D-G}} & \multirow{3}{*}{\makecell{$32\times 2$}} & 
             Baseline &  89.0 & 59.5 & 70.1 & 42.1 & 30.5\\
            & &  + \textbf{FA-ST}  & 90.2 & \textbf{63.6} & \textbf{71.0} & \textbf{44.5} & 31.1 \\
             & &  + \textbf{FA-T}  & \textbf{90.4} & 62.2 & 68.8 & 44.3 & \textbf{31.5} \\
            \hline
        \end{tabular}}
\end{center}
\caption{\textbf{Evaluation results on Mini-Kinetics.} We evaluate FreqAug (FA) with diverse backbones, including SlowOnly-50 (SO-50), SlowOnly-18 (SO-18), R(2+1)D and S3D-G, via finetuning and low-shot learning protocols. Here, $T$: number of frames, $\tau$: input stride. 
}
\label{table:backbone}
\end{table}

\red{ 
In Table~\ref{table:backbone}, we present the evaluation results of MoCo with FreqAug pretrained on MK200. We validate on four different backbones: SlowOnly-50 (SO-50), SlowOnly-18 (SO-18), R(2+1)D, and S3D-G, which have various input resolutions (number of frames $T$, stride $\tau$), depth, and network architecture.
First, MoCo pretrained SO-50 with FreqAug significantly improves the baseline in all five downstream tasks.
The absolute increments of top-1 accuracy range from 2.5\% to 5.3\% depending on the task.
We observe that FreqAug-ST shows comparable or better accuracy than FreqAug-T in four out of five tasks, indicating the synergy between spatial and temporal filters.
The results of the other three backbones show that FreqAug boosts the performance in almost all cases regardless of temporal input resolutions and the network architecture.
Please refer to Appendix~\ref{subsec:ax_ssl} for results with other SSL methods, \ref{subsec:ax_backbone} for the detailed setup of each backbone and results of 3D-ResNet-18 and other input resolutions, and \ref{subsec:ax_other} for comparison with other augmentations.
}

\begin{figure}[t]
\centering
    \includegraphics[width=\linewidth]{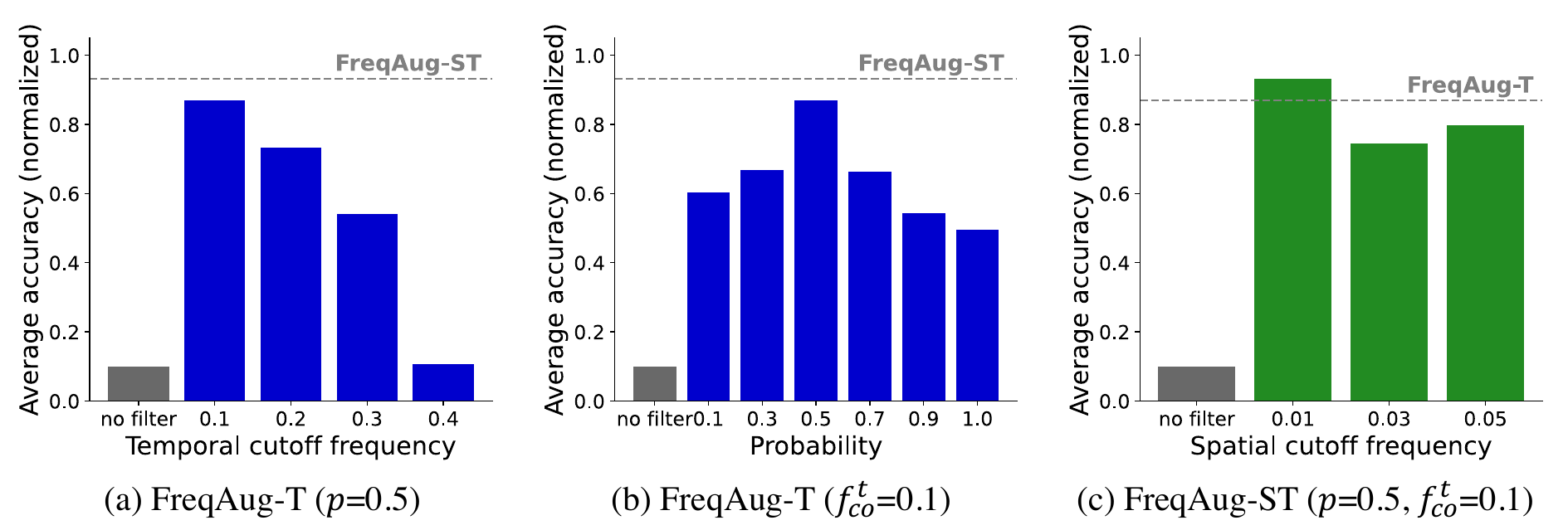}
    \caption{\textbf{Hyperparameter ablations on Mini-Kinetics.} (a) temporal cutoff frequency ($f^t_{co}$) and (b) augmentation probability ($p$) for FreqAug-T, and (c) spatial cutoff frequency ($f^s_{co}$) for FreqAug-ST. Other parameters set fixed with the value in the parenthesis. Min-max normalized accuracies of 5 tasks are averaged.}
\label{fig:ablation}
\end{figure}

\subsection{Ablation Study}
\label{subsec:ablation}
\noindent
\textbf{On Hyperparameters.} 
We conduct three types of ablation studies on MK200 to search for proper hyperparameters in Fig.~\ref{fig:ablation}. 
SO-50 pretrained with MoCo is used as the baseline model.
For ease of visualization, we first min-max normalize top-1 accuracies for each task using all ablation models, then present average accuracy over five action recognition tasks.
We also mark the accuracy of models with default FreqAug-ST or FreqAug-T in dotted line for a better comparison. 
Note that the cutoff frequencies are searched in consideration of the minimum interval between each component: $1/T$ for temporal and $1/H$ (or $1/W$) for spatial dimension.
Fig.~\ref{fig:ablation} shows that FreqAug with default hyperparameters, (a-b) $f^t_{co}{=}0.1$ and $p{=}0.5$ for FreqAug-T, and (c) $f^s_{co}{=}0.01$ for FreqAug-ST, achieves the best performance. 
Detailed description and more ablation studies can be found in Appendix~\ref{subsec:ax_ablation} and~\ref{subsec:ax_view}.

\begin{table}[!t]
\tabcolsep=0.05cm
\begin{center}
\resizebox{0.9\linewidth}{!}{
		\begin{tabular}{c|c|c|c|c|c}
            \hline
            \multirow{2}{*}{\makecell{Filter }} &  \multicolumn{3}{c|}{Finetune} &\multicolumn{2}{c}{Low-shot (10\%)}\\
            \cline{2-6}
            &  UCF101 & HMDB51 & Diving48 & Gym99 & SSv2 \\
            \hline
            No filter & 87.0 & 56.5 & 67.8 & 29.9 & 25.3\\
            \textbf{HPF (default)} &  \textbf{89.8} & \textbf{60.8} & \textbf{70.3} & \textbf{35.2} & \textbf{28.1} \\
            \hline
            LPF ($f^t_{co}$=0.2) &  84.1 & 51.3 & 66.3 & 26.2 & 22.2 \\            
            LPF ($f^t_{co}$=0.3) &  85.8 & 54.4 & 67.9 & 28.8 & 24.2 \\
            LPF ($f^t_{co}$=0.4) &  87.9 & 56.1 & 69.2 & 30.3 & 25.5 \\
            \hline
            Random ($M$=2) &  88.9 & 59.0 & 69.1 & 33.4 & 26.9 \\
            Random ($M$=3) &  89.1 & 58.0 & 69.1 & 33.3 & 25.9 \\
            Random ($M$=5) &  88.2 & 56.5 & 69.5 & 31.5 & 25.2 \\
            \hline
        \end{tabular}}
\end{center}
\caption{\textbf{Temporal filtering strategy comparison on Mini-Kinetics:} 1) LPF with cutoff frequency ($f^t_{co}$) and 2) random masking policy with mask parameter ($M$).}
\label{table:filter}
\end{table}

\noindent
\textbf{On Filtering Strategy.}
\red{ 
In Table~\ref{table:filter}, we compare two variants of temporal filtering strategy on MoCo-pretrained SO-50: LPF and random masking.
LPF strategy is masking frequency components less than $f^t_{co}$ as opposed to default HPF.
We tested $f^t_{co}{\in}\{0.2,0.3,0.4\}$ and observe that the performance becomes worse than the baseline as more high-frequency components are filtered out.
The results show a clear contrast between HPF and LPF strategies, and choosing a proper frequency band for the filter is essential.
We also tested temporal random mask, like SpecAugment~\cite{park2019specaugment}, with mask parameter $M$. Larger $M$ indicates that a larger mask size can be sampled. Refer to Appendix~\ref{subsec:ax_code_random} for the detail.
The scores for random policy ($M{\in}\{2,3,5\}$) are better than the baseline but cannot reach the HPF policy's score, which confirms the validity of selective augmentation.
Refer to Appendix~\ref{subsec:ax_stfilter} for filtering in video domain.
}

\subsection{Comparison with Previous Models}
\begin{table}[!t]
\tabcolsep=0.05cm

\begin{center}
\resizebox{0.95\linewidth}{!}{
		\begin{tabular}{=l|+c|+c|+c|+c|+c|+c}
            \hline
            \multirow{2}{*}{\makecell{Model}} & \multirow{2}{*}{\makecell{\small{Backbone}}} & \multirow{2}{*}{\makecell{T}} & \multirow{2}{*}{\makecell{\small{Epochs}}} & \multicolumn{3}{c}{Finetune}\\
            \cline{5-7}
            & & & & \small{UCF101} & \small{HMDB51} & \small{Diving48} \\
            \hline
            RSPNet\ddag & S3D-G & 64 & 200  & 89.9 & 59.6 & N/A \\
            MoCo-BE & I3D & 16 & 50  & 86.8 & 55.4 & 62.4 \\
            FAME \dag & I3D & 16 & 200  & 88.6 & 61.1 & 72.9 \\
            ASCNet \dag & S3D-G & 64 & 200 & 90.8 & 60.5 & N/A \\
            $\rho$MoCo ($\rho=2$)\dag & SO-50 & 8 & 200  & 91.0 & N/A & N/A \\
            $\rho$BYOL ($\rho=2$)\dag & SO-50 & 8 & 200  & 92.7 & N/A & N/A \\
            \hline
            \rowstyle{\color{gray}}
            CVRL & SO-50 & 32 & 800  & 92.2 & 66.7 & N/A\\
            \rowstyle{\color{gray}}
            RSPNet\ddag & S3D-G & 64 & 1000  & 93.7 & 64.7 & N/A\\
            \rowstyle{\color{gray}}
            $\rho$BYOL ($\rho=4$)\dag & SO-50 & 16 & 800  & 95.5 & 73.6 & N/A\\
            \hline
            MoCo (ours) & SO-50  & 8 & 200 & 90.6 & 62.8 & 72.9 \\
            MoCo + \textbf{FreqAug-ST} & SO-50  & 8 & 200 & 92.1 & 65.6 & 74.0 \\
            MoCo + \textbf{FreqAug-T} & SO-50  & 8 & 200 & 91.8 & 65.1 & 73.8 \\
            \hline
            BYOL (ours) & SO-50  & 8  & 200 & 92.9 & 67.7 & 71.9 \\
            BYOL + \textbf{FreqAug-ST} & SO-50  & 8  & 200 & \textbf{93.7} & \textbf{68.3} & \textbf{74.4} \\
            BYOL + \textbf{FreqAug-T} & SO-50  & 8  & 200 & 93.2 & 67.7 & 72.2 \\            
             \hline
        \end{tabular}}
\end{center}
\caption{\textbf{Comparison with RGB-based models pretrained on Kinetics-400.} 
Backbone, number of frames (T), and pretraining epochs are specified. 
The UCF101 and HMDB51 accuracies are averaged over 3 splits.
Models highlighted in gray are pretrained with larger epochs. \dag: evaluated on split-1; \ddag: ambiguous or not specified which splits are used.
}
\label{table:compare}
\end{table}

\red{ 
Table \ref{table:compare} presents K400 experiments with FreqAug compared to previous video SSL works.
For a fair comparison, SSL models are chosen based on three criteria: augmentation-based, RGB-only (without multimodality including optical flow), and spatial resolution of $224\times 224$.
We report the average accuracy of 3 splits for UCF101 and HMDB51.
We set $p{=}0.3$ for BYOL + FreqAug-ST.
Note that $\rho$ of $\rho$MoCo and $\rho$BYOL indicates the number of views from different timestamps, so models with $\rho{=}2$ are directly comparable to our models.
First, both FreqAug-ST and FreqAug-T consistently outperform the baseline MoCo and BYOL on UCF101, HMDB51, and Diving48.
Compared with other models trained with similar epochs, MoCo and BYOL with FreqAug outperform all the others with similar training epochs.
Interestingly, FreqAug demonstrates its training efficiency by defeating RSPNet on HMBD51 and surpassing CVRL; they are pretrained for 1000 and 800 epochs, respectively. 
We expect training with more powerful SSL methods and longer epochs can be complementary to our approach.
}

\begin{table}[!t]
\tabcolsep=0.1cm

\begin{center}
\resizebox{0.95\linewidth}{!}{
		\begin{tabular}{l|c|c|c|c|c|c}
            \hline
            Method & Pretrain & Acc. & Edit & \multicolumn{3}{c}{F1@\{0.10, 0.25, 0.50\}}\\
            \hline
            SO-50 $\dagger$ & Sup. & 59.0 & 59.5 & \hspace*{1.5mm}54.7\hspace*{1.5mm} & \hspace*{1.5mm}49.2\hspace*{1.5mm} & 37.6\\ 
            LSFD, N & Self-sup. & 60.6 & 60.0 & 52.0 & 42.8 & 35.3\\
            \hline
            MoCo$\dagger$ & \multirow{3}{*}{\makecell{Self-sup.}} & 59.9 & 60.4 & 57.2 & 52.0 & 40.2\\
            \ \ + \textbf{FreqAug-ST}$\dagger$ &  & 65.2 & 63.9 & 61.7 & 56.6 & 45.2 \\
             \ \ + \textbf{FreqAug-T}$\dagger$ & & \textbf{65.9} & \textbf{64.8} & \textbf{62.5} & \textbf{57.1} & \textbf{45.3}  \\
            \hline
        \end{tabular}}
\end{center}
\caption{
\textbf{Temporal action segmentation on Breakfast.} 
All features are evaluated with MS-TCN. 
`Edit' denotes edit distance.
$\dagger$: scores are averaged over 10 evaluations on split-1.
}
\label{table:seg}
\end{table}

\begin{table}[!t]
\tabcolsep=0.1cm

\begin{center}
\resizebox{0.95\linewidth}{!}{
		\begin{tabular}{l|c|c|c|c|c|c|c}
            \hline
            Method & Pretrain & \multicolumn{5}{c|}{mAP@\{0.3, 0.4, 0.5, 0.6, 0.7\}} & Avg\\
            \hline
            TSM & \multirow{2}{*}{\makecell{Sup.}} & \hspace*{0.5mm}46.6\hspace*{0.5mm} & \hspace*{0.5mm}39.5\hspace*{0.5mm} & \hspace*{0.5mm}30.1\hspace*{0.5mm} & \hspace*{0.5mm}20.1\hspace*{0.5mm} & \hspace*{0.5mm}12.2\hspace*{0.5mm} & 29.7 \\ 
            TSM + BSP & & 52.3 & 46.3 & 39.8 & \textbf{30.8} & \textbf{21.1} & 38.1 \\
            \hline
            TSN$\dagger$ & \multirow{2}{*}{\makecell{Sup.}} & 45.7 & 36.8 & 28.2 & 19.0 & 11.3 & 28.2 \\
            SO-50$\dagger$ & & 51.1 & 44.2 & 34.2 & 24.7 & 15.3 & 33.9 \\
            \hline
            MoCo$\dagger$ & \multirow{3}{*}{\makecell{Self-sup.}} & 52.2 & 45.6 & 37.3 & 28.0 & 18.2 & 36.3 \\
             \ \ + \textbf{FreqAug-ST}$\dagger$ &  & 54.1 & 47.4 & 39.4 & 29.6 & 19.8 & 38.1 \\
             \ \ + \textbf{FreqAug-T}$\dagger$ &  & \textbf{55.4} & \textbf{48.7} & \textbf{40.3} & 30.3 & 20.2 & \textbf{39.0} \\
            \hline
        \end{tabular}}
\end{center}
\caption{
\textbf{Temporal action localization on THUMOS'14.} 
Features are pretrained on K400 and evaluated with G-TAD. 
$\dagger$: scores are mean over 5 runs.
}
\label{table:loc}
\end{table}

\subsection{Other Downstream Evaluation Results}
In Table~\ref{table:seg}, we report the results of temporal action segmentation task on the Breakfast dataset.
We experiment with the features extracted from MoCo pretrained SO-50 on K400.
In addition, we report the performance of the extracted feature by officially released SO-50~\cite{fan2020pyslowfast} pretrained on K400 by supervised learning.
The results show that MoCo-pretrained with FreqAug substantially improves the baseline on all metrics.
We conjecture that foreground motion can easily be separated in the videos with static backgrounds by the FreqAug-enhanced feature.
Furthermore, MoCo with FreqAug surpasses its supervised counterpart and LSFD~\cite{Behrmann2021longshortview} in all metrics, which is the only video SSL method evaluated on this task.

In Table~\ref{table:loc}, we report the results of temporal action localization on THUMOS'14 dataset.
We use the features extracted from MoCo pretrained SO-50 on K400.
The results show that MoCo features outperform supervised features from RGB-only TSN~\cite{wang2016temporal} (two-stream model is used in original G-TAD) and SO-50. 
Moreover, adding FreqAug to MoCo improves the localization performance even further than the baseline.
We also compare our results to BSP~\cite{xu2021boundary}, a localization-specific pre-training method, showing similar or better localization performances.
Note that BSP~\cite{xu2021boundary} is pre-trained in supervised manners while our encoders are pre-trained with fully unsupervised.
\red{
For more results and analysis, please refer to Appendix~\ref{subsec:ax_byol} and~\ref{subsec:ax_seg}. 
}

\begin{table}[t]
\tabcolsep=0.03cm

\begin{center}
\resizebox{0.95\linewidth}{!}{
		\begin{tabular}{l|c|c|c|c|c|c}
            \hline
            \multirow{2}{*}{\makecell{Method }} &
            Sup.&
            \multicolumn{3}{c|}{Finetune} &\multicolumn{2}{c}{Low-shot (10\%)}\\
            \cline{2-7}
            & \small{MK200} & \small{UCF101} & \small{HMDB51} & \small{Diving48} & \small{Gym99} & \small{SSv2} \\
            \hline
            SlowOnly-50  & 77.4 & 91.0 & 61.0 & 72.3 & 36.4 & 25.5\\
             + \textbf{FreqAug-ST} & 78.6 & 91.3 & 62.9 & 73.2 & 39.2 & 27.1 \\
            + \textbf{FreqAug-T} & 78.0 & 91.5 & 65.4 & 71.0 & 40.0 & 26.2 \\
            \hline
        \end{tabular}}
\end{center}
\caption{\textbf{Supervised pretraining with FreqAug.} SlowOnly-50 pretrained on MK200. Sup. denotes supervised action recognition accuracy.}
\label{table:super}
\end{table}

\section{Discussion}
\label{sec:disc}

\subsection{FreqAug for Supervised Learning}
One may wonder whether using FreqAug in supervised learning is still effective; here, we evaluate FreqAug in a supervised scenario to demonstrate the versatility of our method. 
Table~\ref{table:super} shows the performance of MK200 pretrained SlowOnly-50 by supervised learning for 250 epochs.
Note that $p{=}0.3$ is used since we observed lower accuracy with a too large $p$.
When we applied FreqAug on top of basic augmentation, we observe overall performance improvements, including the performance of the five downstream tasks and the MK200 pretraining task.

\subsection{Influence on Video Representation Learning}
\label{subsec:tsne}
\begin{figure}[t]
  \centering
   \includegraphics[width=\linewidth]{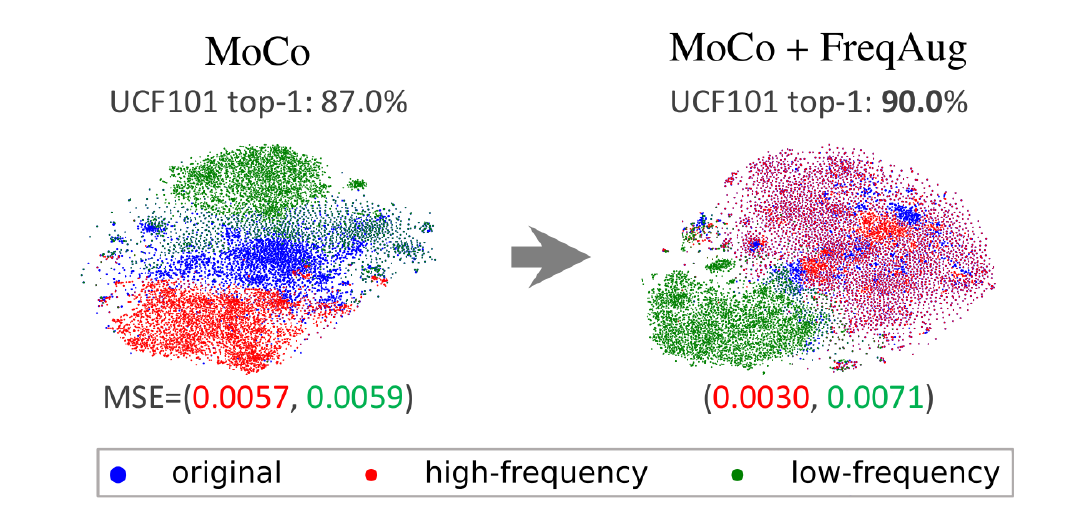}
   \caption{\textbf{t-SNE visualization of the output features from original frames (blue) and its temporal HFC (red) or LFC (green).} Mean squared error (MSE) between original features with HFC/LFC are presented under each plot. MoCo pretrained SlowOnly-50 models with or without FreqAug (and UCF101 finetuning acccuracies) are compared. FreqAug makes features of HFC close to that of original clips which results in better downstream performance. If red and blue dots are too close, they can be perceived as purple.}
   \label{fig:tsne}
\end{figure}
\begin{figure}[!t]
  \centering
   \includegraphics[width=0.95\linewidth]{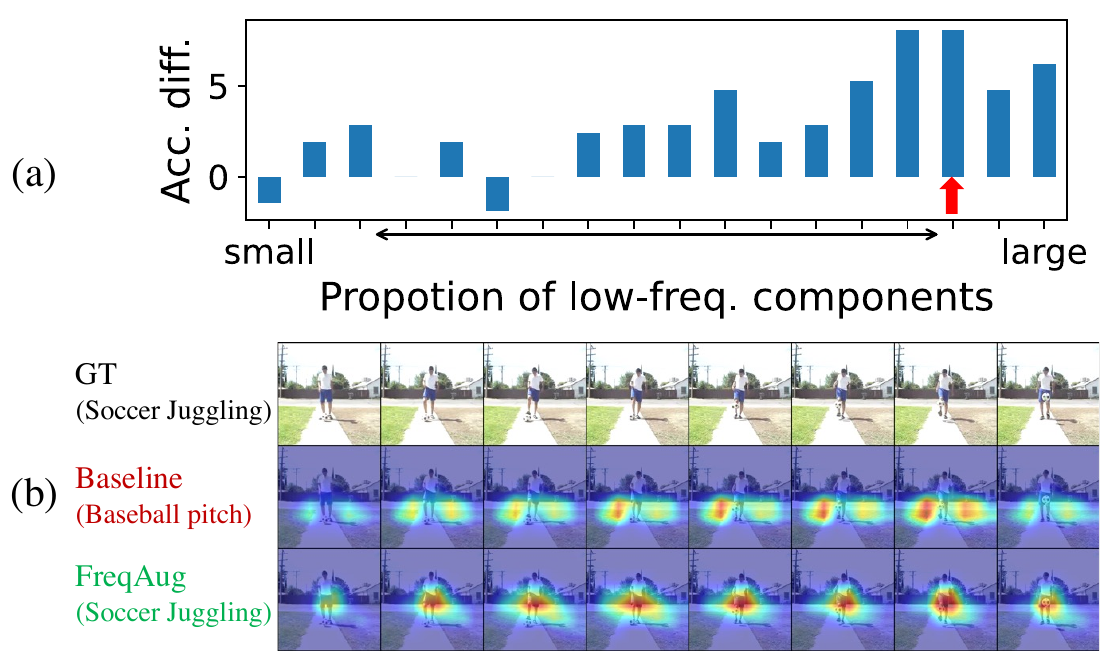}
   \caption{\textbf{Comparing downstream models pretrained with or without FreqAug on UCF101.} (a) Accuracy difference according to the LFC ratio of the sample. (b) Grad-CAM of a clip from a bin with large LFC (red arrow in (a)).}
   \label{fig:gradcam}
\end{figure}
We take a closer look at how the downstream performance of the features learned through FreqAug can be improved compared to the baseline.
Fig. \ref{fig:tsne} shows t-SNE~\cite{Maaten2008tsne} plots of features from original clips (blue) with both high-frequency components (HFC) and low-frequency components (LFC) and either high-pass or low-pass filtered clips (red/green) in temporal dimension.
The distance between two features is measured using mean squared error (MSE).
We compare features from MoCo pretrained SlowOnly-50 on MK200 with or without FreqAug-ST.
The samples are from the validation set of MK200, and $f^t_{co}{=}0.2$ are set for both HPF and LPF.
We observe that the distance between original clips and its temporal HFC substantially decreased when the model is pretrained with FreqAug while there are relatively small changes in the distance between the clip and its LFC; which means FreqAug does not reduce the overall distance between features.
It indicates that FreqAug makes the model extract relatively more features from HFC via invariance learning between HFC and all frequency components in the original signal.
We believe the feature representation learned via FreqAug whose HFC has been enhanced, leads to better transferability of the model as empirically shown in Sec. \ref{sec:exp}.
Refer to Appendix~\ref{subsec:ax_tsne} for more t-SNE analysis on FreqAug.

\red{
To analyze the effect of FreqAug on the downstream task, we group data instances in UCF101 according to the amount of temporal LFC each video has and present accuracy increment in each group caused by FreqAug in Fig.~\ref{fig:gradcam} (a); refer to Sec.~\ref{subsec:ax_group} for the detailed description.
The result shows that the effectiveness of FreqAug tends to be amplified even more on videos with a higher proportion of temporal LFC; those videos are expected to have a large portion of static scenes, background, or objects.
In Fig.~\ref{fig:gradcam}(b), we visualize a sample from a bin with a large LFC (red-arrowed in (a)); original frames, GradCAM~\cite{Selva2017gradcam} of MoCo baseline (Baseline) and MoCo+FreqAug (FreqAug) models from top to bottom.
We observed that FreqAug correctly focuses on the person juggling a soccer ball while Baseline fails to recognize the action because it focuses on the background field. 
Refer to Appendix~\ref{subsec:ax_gradcam} for more samples.
In conclusion, FreqAug helps the model focus on motion-related areas in the videos with static backgrounds.
}

\subsection{\red{Analysis on Temporal Filtering}} 
\begin{figure}[t]
  \centering
   \includegraphics[width=0.95\linewidth]{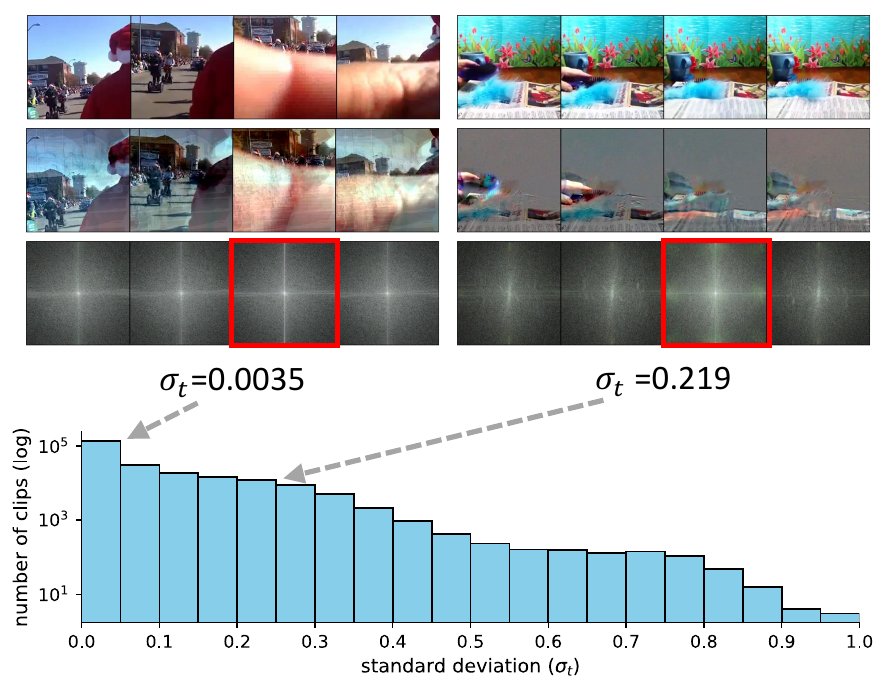}

   \caption{\textbf{Standard deviation of spectrum intensity over temporal axis.} The histogram illustrates the standard deviation (std) distribution of clips in the K400 training set. Top figures show examples of small std (left) and large std (right) videos; top, middle, and bottom rows denote original frames, filtered frames, and spectrum with its std, respectively. Red box indicates where the temporal frequency is zero.}
   \label{fig:discuss}
\end{figure}
\begin{table}[!t]
\tabcolsep=0.05cm
\begin{center}
\resizebox{0.85\linewidth}{!}{
		\begin{tabular}{c c|c|c|c|c|c}
            \hline
            \multicolumn{2}{c|}{Std.} & \multicolumn{3}{c|}{Finetune} &\multicolumn{2}{c}{Low-shot(10\%)}\\
            \cline{1-7}
            \multicolumn{1}{c|}{\small{$<0.05$}} & \small{$>0.05$} & \small{UCF101} & \small{HMDB51} & \small{Diving48} & \small{Gym99} & \small{SSv2} \\
            \hline
            \multicolumn{2}{c|}{no FreqAug} & 87.0 & 56.5 & 67.8 & 29.9 & 25.3\\
            \hline
            &\checkmark & 89.4 & 59.3& \textbf{70.9} & 32.4 & 27.4 \\
            \checkmark & & 88.8 & 58.3 & 69.9 & 32.4 & 27.0 \\
            \hline
            \checkmark & \checkmark & \textbf{89.8} & \textbf{60.8} & 70.3 & \textbf{35.2} & \textbf{28.1} \\
            \hline
        \end{tabular}}
\end{center}
\caption{\textbf{Impact of samples with large temporal variations to the temporal filter.} Samples of temporal spectrum std. below or above threshold (0.05) are rejected to apply the temporal filter in FreqAug-T. Tested on MK200. }
\label{table:minus}
\end{table}


As aforementioned, FreqAug can help the model focus on motion-related information by randomly removing the background with a temporal high-pass filter. However, one may doubt whether FreqAug is only effective with videos whose background can be easily removed.
In order to resolve the doubt above, we conduct further analysis by applying FreqAug on different subsets of the training dataset according to the spectrum intensity over the temporal axis.

As in the top left clip of Fig.~\ref{fig:discuss}, videos with a low standard deviation of spectrum intensity over the temporal frequency ($\sigma_t$) tend to have temporally varying backgrounds due to rapid camera moving or abrupt scene change, which makes a naive filter hard to remove background.
The spectrum intensity will be concentrated on the temporal zero-frequency (red boxes) when the scene change is small over time (right).
Otherwise, the spectrum spreads across all temporal frequencies (left).
In other words, $\sigma_t$ gets decreased if many scene transitions exist.
For quantitative analysis, we take the logarithm and mean over spatial frequency to the spectrum and then calculate std. over time. 
As we expected, the background of videos with a small $\sigma_t$ is often not eliminated, and some traces of other frames are mixed.

The histogram in Fig.~\ref{fig:discuss} shows that around half of the clips in K400 have relatively small $\sigma_t$ (below 0.05)
Then, a question naturally arises about how those clips with small $\sigma_t$ affect the learning with FreqAug.
We argue that the clips with a visually remaining background are also helpful for FreqAug.
To support our claim, we conduct a quantitative experiment in Table~\ref{table:minus} to confirm the impact of the temporal filter on videos with small $\sigma_t$ (0.05).
We study with two variants of FreqAug-T, which exclude the clips of either small $\sigma_t$ (\ie, under 0.05) or large $\sigma_t$ (\ie, over 0.05) when applying the filter.
The result demonstrates that FreqAug outperforms the baseline with a large margin in every case, even in the case of clips with small $\sigma_t$.
This implies that the temporal filter enhances the representation of clips with both small and large temporal variations.
Therefore, this experiment validates our claim that the role of temporal filtering is not limited to background erasing.

\section{Conclusion}
\label{sec:conc}
In this paper, we have proposed a simple and effective frequency domain augmentation for video representation learning.
FreqAug augments multiple views by randomly removing spatial and temporal low-frequency components from videos so that a model can learn from the essential features.   
Extensive experiments have shown the effectiveness of FreqAug for various self-supervised learning frameworks and diverse backbones on \textit{seven} downstream tasks.
\red{Lastly, we analyze the influence of FreqAug on both video SSL and its downstream tasks.}

\section*{Acknowledgements}
This work was supported by NAVER Corporation. 
The computational work in this study was mostly conducted on NAVER Smart Machine Learning (NSML) platform~\cite{nsml1,nsml2}.

\bibliography{main}

\clearpage
\section*{Appendix}
\appendix
\setcounter{table}{0}
\renewcommand{\thetable}{A\arabic{table}}
\setcounter{figure}{0}
\renewcommand{\thefigure}{A\arabic{figure}}
\setcounter{listing}{0}
\renewcommand\thelisting{A\arabic{listing}}
\renewcommand\thesection{A\arabic{section}}
\section{Datasets and Implementation Details}
\label{sec:ax_implementation}
\subsection{Datasets.}
{\noindent \bf Pretrain.}
For pretraining the model, we use Kinetics-400 (K400)~\cite{Carreira2017QuoVA} and Mini-Kinetics (MK200)~\cite{Xie2018rethinking}. 
K400 is a large-scale video action recognition dataset with 400 classes, and MK200 is a balanced subset of K400 with 200 classes.
With the limited resources, K400 is too large for extensive experiments, so we choose MK200 as a major testbed to verify our method's effectiveness.
We also report results on K400 for comparison to previous models.

{\noindent \bf Action Recognition.}
For evaluation of the pretrained models, we use five different action recognition datasets: UCF101~\cite{soomro2012ucf101}, HMDB51~\cite{kuehne2011hmdb}, Diving48~\cite{Li2018RESOUND}, Gym99~\cite{shao2020finegym}, and Something-something-v2 (SSv2)~\cite{Goyal2017something}.
Following the standard practice, we report the accuracy of the finetuned model on the three datasets: UCF101, HMDB51, and Diving48.
Note that we present split-1 accuracy for UCF101 and HMDB51 by default unless otherwise specified.
For Gym99 and SSv2, we evaluate the models on the low-shot learning protocol using only 10\% of training data.
Finetuning on those datasets may dilute the effect of pretraining since they are relatively large-scale (especially the number of samples in SSv2 is about twice larger than that of our main testbed MK200).
Nevertheless, evaluating models on those datasets is still meaningful because it reflects the impact of pretraining on a substantially different target domain. 
Furthermore, low-shot learning can simulate practical scenarios where only a few labels are available for the target task.

{\noindent \bf Localization.}
Breakfast~\cite{Kuehne12breakfast} dataset contains 1,712 untrimmed videos with temporal annotations of 48 actions related to breakfast preparation.
On average, six action instances are contained in each video.
THUMOS’14~\cite{idrees2017thumos} is a dataset that consists of 413 untrimmed videos of 20 actions for the temporal action localization task; 200 videos for the training and 213 videos for the evaluation.

\subsection{Self-supervised Pretraining}
\begin{table*}[!t]
\tabcolsep=0.05cm
\begin{center}
\resizebox{0.7\linewidth}{!}{
		\begin{tabular}{l|c|c|c|c|c|c|c}
            \hline
            \bf SSL Method & \multicolumn{2}{c|}{\bf MoCo} & \multicolumn{2}{c|}{\bf BYOL} &\bf SimSiam &\bf SimCLR &\bf SwAV \\
            \hline
            \bf Dataset &\bf MK200 & \hspace*{0.7mm}\bf K400\hspace*{0.7mm} &\bf MK200 & \hspace*{0.7mm}\bf K400\hspace*{0.7mm} &\bf MK200 &\bf MK200 &\bf MK200 \\
            \hline
            Epochs & \multicolumn{7}{c}{200 (linear warm-up for 35 epochs)} \\
            Batch size & 128 & 256 & 64 & 256 & 64 & 128 & 128 \\
            Learning rate & 1.6 & 4.0 & 1.6 & 4.8 & 1.6 & 3.2 & 3.2 \\
            Weight decay & 1e-5 & 1e-5 & 1e-5 & 1e-6 & 1e-5 & 1e-5 & 1e-5\\
            Optimizer & SGD & SGD & SGD & LARS & SGD & SGD & SGD\\
            Number of GPUs & 4 & 8 & 4 & 8 & 4 & 4 & 4 \\            
            \hline             
            InfoNCE, temperature & \multicolumn{2}{c|}{0.1} & \multicolumn{2}{c|}{-} & - & 0.1 & 0.1\\ 
            Dictionary size & \multicolumn{2}{c|}{65536} & \multicolumn{2}{c|}{-} & - & - & - \\
            Momentum coefficient & \multicolumn{2}{c|}{0.994} & \multicolumn{2}{c|}{0.99 } & -  & - & - \\
            BatchNorm & \multicolumn{2}{c|}{Shuffle BN} & \multicolumn{2}{c|}{SyncBN } & SyncBN  & SyncBN & SyncBN \\
            MLP, BatchNorm & \multicolumn{2}{c|}{-} & \multicolumn{2}{c|}{\checkmark } & \checkmark  & \checkmark & \checkmark \\ 
            MLP, output dim. & \multicolumn{2}{c|}{128} & \multicolumn{2}{c|}{256} & 2048 & 2048 & 2048\\
            Projection MLP, layers & \multicolumn{2}{c|}{2} &\multicolumn{2}{c|}{2}& 3 & 3 & 3\\
            Projection MLP, hidden dim. & \multicolumn{2}{c|}{2048} & \multicolumn{2}{c|}{4096} & 2048 & 2048 & 2048\\            
            Prediction MLP, layers & \multicolumn{2}{c|}{-} &\multicolumn{2}{c|}{2}& 2 & - & - \\
            Prediction MLP, hidden dim. & \multicolumn{2}{c|}{-} & \multicolumn{2}{c|}{4096} & 512 & - & - \\
            \hline
        \end{tabular}}
\end{center}

\caption{\textbf{Pre-training configurations.}}
\label{table:ax_setting}
\end{table*}
For self-supervised pretraining, we use SGD with a momentum of 0.9, a half-cosine learning rate schedule~\cite{Loshchilov2017SGDR}, and a linear warm-up for the first 35 epochs.
By default, all the models are trained for 200 epochs.
The base learning rate is set to $1.6$, and weight decay is set to 1e-5 as a default.
The linear warm-up starts from 0.001.
We use LARS~\cite{You2017lars} optimizer with a trust coefficient of 0.001 for BYOL on K400.
Details for each SSL method are presented in Table~\ref{table:ax_setting}.

In terms of spatial augmentation, a combination of resizing, cropping, horizontal flipping, color jittering, Gaussian blurring, and random grayscaling is applied in a temporally consistent manner as our baseline following~\cite{Chen2020mocov2}. 
For random resized cropping, we randomly sample a sub-region with the scale between (0.2, 1.0) and the aspect ratio between (3/4, 4/3) of the original frame, followed by scaling the size to $224\times224$.
For color jittering, brightness, contrast, saturation, and hue are adjusted randomly with a probability of 0.8.
Each attribute is modified by using {\fontfamily{qcr}\selectfont adjust\_\{attribute\}} method in {\fontfamily{qcr}\selectfont torchvision.transforms.functional} module of Pytorch~\cite{NEURIPS2019_9015}.
Each factor is uniformly sampled within a range of [0.6, 1.4] for the first three attributes and a range of [-0.1, 0.1] for the last.
For BYOL experiment on K400, all factors are used at half the value.
Gaussian blurring is applied with the probability of 0.5 and the standard deviation sampled from $Uniform(0.1, 2.0)$.
Frames are turned into grayscale with the probability of 0.2.

For temporal augmentation, randomly sampled clips from different timestamps compose the positive instances for learning temporally persistent representation.
Also, we put a constraint on the temporal sampling that two clips should be sampled within a range of 1 second from each clip’s center frame.
Each clip consists of $T$ frames sampled from $T\times \tau$ consecutive frames with stride $\tau$.
We apply the identical baseline augmentation strategy for all the SSL methods used in the main paper.

Recently, some deep learning libraries, \eg ,Pytorch~\cite{NEURIPS2019_9015}, offer a GPU accelerated version of fast Fourier transform (FFT) and inverse FFT (iFFT) functions, which are efficient implementations of DFT and iDFT. Thus we utilize these algorithms for FreqAug.
We choose two default settings for FreqAug unless specified.
FreqAug-T (temporal) uses a temporal high-pass filter (HPF) with a cutoff frequency of 0.1 and a probability of 0.5.
FreqAug-ST (spatio-temporal) is a combination of spatial HPF with a cutoff frequency of 0.01 alongside FreqAug-T.

\subsection{Downstream Tasks}
\noindent
\textbf{Finetuning and low-shot learning.}
For supervised finetuning and low-shot learning, we train the models for 200 epochs with a base learning rate of 0.025, zero weight decay, and without warm-up.
Only basic spatial augmentations are used for finetuning and low-shot learning: random resizing (shorter side to [256, 320]), random cropping (224$\times$224 pixels from resized frames), and random horizontal flipping. 
Random horizontal flipping is excluded in the case of SSv2 because the dataset has direction-sensitive action categories.
For UCF101~\cite{soomro2012ucf101} and HMDB51~\cite{kuehne2011hmdb}, temporal augmentation draws $T\times \tau$ consecutive frames randomly and uniformly sub-samples $T$ frames for the input.
For Diving48~\cite{Li2018RESOUND}, Gym99~\cite{shao2020finegym}, and Something-something-v2 (SSv2)~\cite{Goyal2017something}, segment-based sampling~\cite{wang2016temporal} is applied, which divides a video into $T$ equal length segments and samples one frame randomly from each segment.
Dropout with the drop probability of 0.8 is applied before the linear classifier.

\noindent
\textbf{Temporal Action Localizaton.}
We evaluate the learned representation on two temporal action localization tasks: action segmentation which allocates every frame into pre-defined action classes, and action localization which localizes class-specific temporal action snippets from backgrounds.
Both tasks aim to evaluate the ability of learned features to recognize temporal action changes.

We train an action segmentation model, Multi-Stage Temporal Convolutional Network (MS-TCN)~\cite{Farha2019mstcn}, on Breakfast dataset~\cite{Kuehne12breakfast} with frozen features from the pretrained encoder.
MS-TCN is composed of four stages with ten layers each.
Each layer is a sequence of a 1D dilated convolution, ReLU activation, and a $1\times1$ convolution and has a residual connection.
The features of videos in Breakfast dataset are extracted by the pretrained encoder with the sliding window fashion.
The FPS of the videos is fixed to 15, and sliding window size is $21\times1$ (number of frames $\times$ stride).
Refer to~\cite{Farha2019mstcn} for other MS-TCN settings.  

For the temporal action localization task, we train Graph-Temporal Action Detection (G-TAD)~\cite{xu2020g} model on THUMOS'14~\cite{idrees2017thumos} dataset.
G-TAD treats a video sequence as a graph and solves action detection as a sub-graph localization problem using a set of graph convolution network (GCN)~\cite{Kipf2017gcn} and a sub-graph of interest alignment layer.
We extract video features with a $9\times1$ sliding window in the case of the pretrained SlowOnly encoder; we use officially released feature~\footnote{https://github.com/frostinassiky/gtad} for TSN~\cite{wang2016temporal}.
All other G-TAD related settings are set fixed as its original paper~\cite{xu2020g}.

\subsection{Evaluation}
We report top-1 accuracy as the performance measure for action recognition downstream tasks.
For Kinetics, UCF101, and HMDB51, we report the averaged accuracy over 30-crops, \ie, temporally 10-crops and spatially 3-crops, following standard practices~\cite{Chris2019slowfast}.
All frames are resized to fit its shorter side to 256, and three 256$\times$256 crops are uniformly sampled along the other axis.
Similarly, ten clips, equally distributed in the temporal axis, are sampled for each spatial crop.
For Diving48, Gym99, and SSv2, we report spatially 3-crop accuracy with the segment-based temporal sampling.
Note that $\tau$ can be ignored for the datasets with the segment-based sampling strategy.

Following~\cite{Farha2019mstcn}, the evaluation metrics for the temporal action segmentation task are the frame-wise accuracy, the edit distance, and the F1 scores at the overlapping threshold of 10\%, 25\%, and 50\%.
The temporal intersection over union (IoU) measures the overlap between the predicted action and the ground truth for the F1 score.
The models are evaluated on the split-1 of Breakfast dataset, following the previous work~\cite{Behrmann2021longshortview}.
For better reliability, we report average scores of 10 evaluations with different random seeds.

For temporal action localization, we measure mean average precision (mAP) at IoU threshold from 0.3 to 0.7 with stride 0.1 following~\cite{xu2020g}.
We also report average mAP over the five IoU thresholds.
All metrics are averaged over five different runs for better reliability.

\subsection{Models}
\noindent
\textbf{Backbone.}
Our default encoder backbone is SlowOnly-50, a variant of 3D Residual Network (ResNet) that originated from the slow branch of SlowFast network~\cite{Chris2019slowfast}.
It shows decent action recognition performances with affordable computations compared to other variants thanks to its architectural choices, \eg, decomposed spatial and temporal convolutions, no temporal downsampling, large temporal stride ($\tau$), and temporal convolutions only at the last two stages.  
We also evaluate our method on 3D-ResNet-18 (R-18)~\cite{hara3dcnns}, R(2+1)D~\cite{Tran2018r21d}, and S3D-G~\cite{Xie2018rethinking} models.
R-18 is a 3D-ResNet with full 3D convolutions, and R(2+1)D is another popular variant of 3D ResNet with factorized 3D convolution filter.
S3D-G is an inception-style 3D model with gating modules. 

\noindent
\textbf{SSL Methods.}
We implement MoCo~\cite{Chen2020mocov2}, BYOL~\cite{Jean2020byol}, Simsiam~\cite{Chen2021simsiam}, SimCLR~\cite{Chen2020simclr}, and SwAV~\cite{Caron2020swav} with some hyperparameter changes for the video model (adopt some settings from~\cite{Feichtenhofer2021largescale}).
The configuration of each method is summarized in Table~\ref{table:ax_setting}.
For MoCo and BYOL, we use cosine annealing of the momentum coefficient following~\cite{Feichtenhofer2021largescale}.
Projection and prediction MLP hidden layers are equipped with batch normalization (BN)~\cite{Ioffe2015BN} and a ReLU activation.
The projection MLP of SimSiam has a BN after linear output layer.

\begin{algorithm*}[!t]
\definecolor{codeblue}{rgb}{0.25,0.5,0.5}
\definecolor{codekw}{rgb}{0.85, 0.18, 0.50}
\begin{lstlisting}[language=python]
# x: video, B x C x T x H x W
# p: filter probability
# co_s, co_t: spatial and temporal cutoff freq.
# type_s, type_t: HPF if True else LPF
# fftfreq: calculate FFT sample frequencies
# outer: outer product

def freqaug(x, p, co_s, co_t, type_s, type_t):
    if random() < p: # U(0,1)
        # pass bands
        pass_t = torch.abs(torch.fft.fftfreq(T)) < co_t 
        pass_h = torch.abs(torch.fft.fftfreq(H)) < co_s
        pass_w = torch.abs(torch.fft.fftfreq(W)) < co_s
        
        pass_hw = torch.outer(pass_h, pass_w)
        if type_s:
            pass_hw = torch.logical_not(pass_hw)
            
        if type_t:
            pass_t = torch.logical_not(pass_t)
        
        F = torch.outer(pass_t, pass_hw.view(-1)).view(T, H, W) # filter
        X = torch.fft.fft(x) # FFT, X: spectrum
        X_hat = F * X # filtering
        return torch.fft.ifft(X_hat) # inverse FFT
    else:
        return x
\end{lstlisting}        
\caption{FreqAug: Pytorch-like Pseudocode}
\label{alg:code}
\end{algorithm*}

\section{Pseudo-Code}
\label{sec:ax_code}
\subsection{FreqAug}
\label{subsec:ax_code_fa}
We provide a Python-style pseudo-code of FreqAug with PyTorch~\cite{NEURIPS2019_9015} package in Algorithm~\ref{alg:code}.
The pseudo-code, FreqAug can be implemented with only a few lines.
In the case when the augmentation is applied ({\fontfamily{qcr}\selectfont random()<p}), the filter ({\fontfamily{qcr}\selectfont F}) is constructed according to the cutoff frequencies ({\fontfamily{qcr}\selectfont co\_s, co\_t}) at first, then 1) transforming the input ({\fontfamily{qcr}\selectfont x}) to the spectrum ({\fontfamily{qcr}\selectfont X}) by FFT, 2) applying the filter by element-wise multiplication and 3) transforming the filtered spectrum ({\fontfamily{qcr}\selectfont X\_hat}) back to the original domain by iFFT are sequentially applied.

Since our method is a data augmentation applied before the video sample is fed to the encoder, the attached code is easily integrated into general PyTorch-based video SSL implementations.

\begin{algorithm*}[!t]
\definecolor{codeblue}{rgb}{0.25,0.5,0.5}
\definecolor{codekw}{rgb}{0.85, 0.18, 0.50}
\begin{lstlisting}[language=python]
# x: video, B x C x T x H x W
# M: mask parameter
# fftfreq: calculate FFT sample frequencies
# outer: outer product

def random_masking(x, M):
    ft = 1.0 / T
    freq_t = [float(i) * ft for i in range(0, T // 2 + 1)]
    
    band_s = random.choice(freq_t) # choose start point of reject band
    freq_t_end = [f for f in freq_t if band_s + M*ft > f >= band_s]
    band_e = random.choice(freq_t_end) # choose end point of reject band
    
    pass_h = torch.abs(torch.fft.fftfreq(H)) <= 0.5 # pass all spatial freq.
    pass_w = torch.abs(torch.fft.fftfreq(W)) <= 0.5 # pass all spatial freq.
    pass_t_s = torch.abs(torch.fft.fftfreq(T)) >= band_s
    pass_t_e = torch.abs(torch.fft.fftfreq(T)) < band_e

    pass_hw = torch.outer(pass_h, pass_w)
    pass_t = torch.logical_not(pass_t_s * pass_t_e)

    F = torch.outer(pass_t, pass_hw.view(-1)).view(T, H, W)
    X = torch.fft.fft(x) # FFT, X: spectrum
    X_hat = F * X # filtering
    return torch.fft.ifft(X_hat) # inverse FFT

\end{lstlisting}
\caption{Temporal Random Masking: Pytorch-like Pseudocode}
\label{alg:code_rand}
\end{algorithm*}

\subsection{Temporal Random Filtering Strategy}
\label{subsec:ax_code_random}
We provide a Python-style pseudo-code of random filtering strategy (Table~\ref{table:filter}) with PyTorch~\cite{NEURIPS2019_9015} package in Algorithm~\ref{alg:code_rand}.
This strategy differs from FreqAug in how the filter ({\fontfamily{qcr}\selectfont F}) is made.
While FreqAug selectively filters a specific frequency band, random strategy can filter out any frequency band with the mask size determined by mask parameter ({\fontfamily{qcr}\selectfont M}).
Specifically, the start point of the reject band ($f_{start}$) is sampled from all possible frequency ranges, followed by choosing the endpoint ($f_{end}$) via sampling the mask size $f_m$ from $[1, ... , M/T)$, where T is the number of frames of the input clip; $f_{end}=f_{start}+f_m$. 
Then, the random temporal filter can be described as, 
\begin{equation}
  F^t_{random}[k] = 
    \begin{cases}
        0 & \text{if $f_{start}<=|k|<f_{end}$}\\
        1 & \text{otherwise}.
    \end{cases}
  \label{eq:filter1d_random}
\end{equation}
Note that $f_{start}$ and $f_{end}$ are {\fontfamily{qcr}\selectfont band\_s} and {\fontfamily{qcr}\selectfont band\_e} in the pseudo-code, respectively.

\section{Additional Experiments}
\label{sec:ax_result}
\begin{table*}[!t]
\begin{center}
\resizebox{0.8\linewidth}{!}{
		\begin{tabular}{c|l|c|c|c|c|c}
            \hline
            \multirow{2}{*}{\makecell{Method }} & \multirow{2}{*}{\makecell{Augmentation }} &  \multicolumn{3}{c|}{Finetune} &\multicolumn{2}{c}{Low-shot (10\%)}\\
            \cline{3-7}
            & &  UCF101 & HMDB51 & Diving48 & Gym99 & SSv2 \\
            \hline
            \multirow{3}{*}{\makecell{MoCo~\cite{Chen2020mocov2}}} & Baseline & 87.0 & 56.5 & 67.8 & 29.9 & 25.3\\
             & + \textbf{FreqAug-ST} &  90.0 {(+3.0)}& 61.6 {(+5.1)}& 71.0 {(+3.2)}& 34.8 {(+4.9)}& 28.1 {(+2.8)}\\
             & + \textbf{FreqAug-T} &  89.8 {(+2.8)}& 60.8 {(+4.3)}& 70.3 {(+2.5)}& 35.2 {(+5.3)}& 28.1 {(+2.8)}\\
            \hline
             \multirow{3}{*}{\makecell{BYOL~\cite{Jean2020byol}}} & Baseline & 88.4 & 59.8 & 70.3 & 38.7 & 29.5 \\
             & + \textbf{FreqAug-ST} & 90.5 {(+2.1)} & 63.2 {(+3.4)}& 72.4 {(+2.1)}& 40.7 {(+2.0)}& 31.2 {(+1.7)} \\
             & + \textbf{FreqAug-T} & 90.5 {(+2.1)} & 62.7 {(+2.9)}& 72.5 {(+2.2)}& 40.8 {(+2.1)}& 31.6 {(+2.1)} \\
            \hline
             \multirow{3}{*}{\makecell{SimSiam~\cite{Chen2021simsiam}}} & Baseline & 86.1 & 57.5 & 67.4 & 33.2 & 27.5 \\
             & + \textbf{FreqAug-ST} & 87.3 {(+1.2)} & 57.3 {(-0.2)} & 70.7 {(+3.3)}& 33.7 {(+0.5)}& 28.6 {(+1.1)} \\
             & + \textbf{FreqAug-T} & 86.0 {(-0.1)} & 58.6 {(+1.1)}& 70.4 {(+3.0)}& 34.1 {(+0.9)}& 29.0 {(+1.5)} \\
            \hline
            \multirow{3}{*}{\makecell{SimCLR~\cite{Chen2020simclr}}} & Baseline & 84.1 & 51.9 & 67.2 & 30.4 & 26.1 \\
             & + \textbf{FreqAug-ST} & 86.4 {(+2.3)} & 56.7 {(+4.8)} & 70.0 {(+2.8)}& 34.4 {(+4.0)}& 28.2 {(+2.1)} \\
             & + \textbf{FreqAug-T} & 86.4 {(+2.3)} & 57.1 {(+5.2)}& 68.9 {(+2.7)}& 32.8 {(+2.4)}& 27.8 {(+1.7)} \\
            \hline
            \multirow{3}{*}{\makecell{SwAV~\cite{Caron2020swav}}} & Baseline & 81.5 & 49.5 & 64.8 & 27.5 & 24.2 \\
             & + \textbf{FreqAug-ST} & 83.1 {(+1.6)} & 51.0 {(+1.5)} & 66.8 {(+2.0)}& 29.8 {(+2.3)}& 25.7 {(+1.5)} \\
             & + \textbf{FreqAug-T} & 83.3 {(+1.8)} & 50.8 {(+1.3)}& 66.0 {(+1.2)}&  30.1{(+2.6)}& 25.6 {(+1.4)} \\
            \hline
        \end{tabular}}
\end{center}
\caption{\textbf{Evaluation results of SSL Methods on Mini-Kinetics.} We demonstrate FreqAug with MoCo, BYOL, SimSiam, SimCLR, and SwAV. Relative increments over the baseline augmentation are stated in the parenthesis. Low-shot denotes finetuning with only 10\% of total training samples.}
\label{table:ax_ssl}
\end{table*}

\subsection{Additional SSL Methods}
\label{subsec:ax_ssl}
In addition to MoCo, we evaluate FreqAug on four other SSL methods, BYOL, SimSiam, SimCLR, and SwAV, in Table~\ref{table:ax_ssl}.
Leveraging FreqAug to BYOL also shows improved performance over the baseline augmentation on all downstream tasks. 
The absolute increments of BYOL (1.7\% to 3.4\%) are a bit less than those of MoCo because BYOL has a stronger baseline.
We also demonstrate FreqAug on SimSiam, which has no momentum encoder, unlike MoCo and BYOL.
Note that we set $p=0.1$ and $f^t_{co}=0.2$ for FreqAug-ST for SimSiam. 
FreqAug with SimSiam enhances the accuracy on most downstream tasks, but the overall increase is a bit less than the numbers of other methods. 
We observe the marginally decreased numbers either on UCF101 or HMDB51; we speculate that the lack of the momentum encoder in SimSiam incurs the accuracy degradation. 
One possible explanation is that the momentum encoder in MoCo and BYOL contributes to stabilizing the invariance learning with relatively hard augmentation like FreqAug.
The negative impact of hard augmentations on SimSiam is recently studied in the literature~\cite{bai2021directional}.
In the case of SimCLR and SwAV, FreqAug also consistently improves all downstream tasks.
Note that SwAV shows lower baseline performance as in~\cite{Feichtenhofer2021largescale}, probably because multi-crop augmentation is not applied for a fair comparison.
In conclusion, we show the effectiveness of FreqAug on five different SSL methods.

\begin{table*}[!t]
\tabcolsep=0.13cm
\begin{center}
\resizebox{0.8\linewidth}{!}{
		\begin{tabular}{c|c|c|l||c|c|c|c|c}
            \hline
            \multirow{2}{*}{\makecell{Backbone }} & \multirow{2}{*}{\makecell{$T\times \tau$ }} &
            \multirow{2}{*}{\makecell{$H\times W$ }} & \multirow{2}{*}{\makecell{Augmentation }} & \multicolumn{3}{c|}{Finetune} &\multicolumn{2}{c}{Low-shot (10\%)}\\
            \cline{5-9}
            & &  & & UCF101 & HMDB51 & Diving48 & Gym99 & SSv2 \\
            \hline
            \multirow{3}{*}{\makecell{SO-50}} & \multirow{3}{*}{\makecell{$8\times 8$}} & 
            \multirow{3}{*}{\makecell{$128\times 128$}} &
            Baseline & 82.2 & 53.9 & 64.2 & 30.8 & 25.4\\
              & & & + \textbf{FreqAug-ST} & 85.4 {(+3.2)} & 58.6 {(+4.7)} & 67.3 {(+3.1)} & 33.7 {(+2.9)} & 29.1 {(+3.7)}\\
              & & & + \textbf{FreqAug-T} & 86.0 {(+3.8)}& 58.0 {(+4.1)}& 65.9 {(+1.7)}& 34.3 {(+3.5)}& 28.6 {(+3.2)}\\
            \cline{1-9}
              \multirow{3}{*}{\makecell{SO-18}} & \multirow{3}{*}{\makecell{$8\times 8$}} &
              \multirow{3}{*}{\makecell{$224\times 224$}} &
              Baseline & 82.4 & 51.9 & 65.0 & 26.9 & 21.6 \\
             & & & + \textbf{FreqAug-ST} & 86.7 {(+4.3)} & 55.6 {(+3.7)} & 67.1 {(+2.1)} & 31.0 {(+4.1)} & 24.1 {(+2.5)} \\
             & & & + \textbf{FreqAug-T} & 85.7 {(+3.3)} & 57.1 {(+5.2)}& 68.4 {(+3.4)}& 29.4 {(+2.5)}& 23.9 {(+2.3)} \\
            \hline
              \multirow{3}{*}{\makecell{SO-18}} & \multirow{3}{*}{\makecell{$16\times 4$}} &
              \multirow{3}{*}{\makecell{$224\times 224$}} &
              Baseline & 84.5 & 55.2 & 74.9 & 30.3 & 23.9 \\
             & & & + \textbf{FreqAug-ST} & 88.5 {(+4.0)} & 57.8 {(+2.6)} & 75.8 {(+0.9)} & 35.3 {(+5.0)} & 25.7 {(+1.8)} \\
             & & & + \textbf{FreqAug-T} & 88.7 {(+4.2)} & 58.8 {(+3.6)}& 75.7 {(+0.8)}& 34.7 {(+4.4)}& 26.1 {(+2.2)} \\
            \hline
             \multirow{3}{*}{\makecell{R-18}} & \multirow{3}{*}{\makecell{$16\times 2$}} &
             \multirow{3}{*}{\makecell{$224\times 224$}} &
             Baseline  & 82.6 & 49.5 & 36.6 & 28.0 & 18.9 \\
              & & & + \textbf{FreqAug-ST}  & 86.8 {(+4.2)} & 56.9 {(+7.4)} & 41.0 {(+4.4)} & 32.8 {(+4.8)} & 20.9 {(+2.0)} \\
              & & & + \textbf{FreqAug-T}  & 86.4 {(+3.8)} & 55.9 {(+6.4)}& 39.1 {(+2.5)}& 32.3 {(+4.3)}& 20.8 {(+1.9)} \\
            \hline
              \multirow{3}{*}{\makecell{R(2+1)D}} & \multirow{3}{*}{\makecell{$32\times 2$}} &
              \multirow{3}{*}{\makecell{$128\times 128$}} &
              Baseline & 86.2 & 60.4 & 64.6 & 42.5 & 29.2 \\
             & & & + \textbf{FreqAug-ST} & 90.0 {(+3.8)} & 65.9 {(+5.5)}& 67.7 {(+3.1)}& 48.4 {(+5.9)}& 31.5 {(+2.3)}\\
             & & & + \textbf{FreqAug-T} & 89.5 {(+3.3)} & 65.2 {(+4.8)} & 70.2 {(+5.6)} & 48.3 {(+5.8)} & 30.5 {(+1.3)} \\
            \hline
             \multirow{3}{*}{\makecell{S3D-G}} & \multirow{3}{*}{\makecell{$32\times 2$}} & 
             \multirow{3}{*}{\makecell{$224\times 224$}} &
             Baseline &  89.0 & 59.5 & 70.1 & 42.1 & 30.5\\
            & & & + \textbf{FreqAug-ST}  & 90.2 {(+1.2)} & 63.6 {(+4.1)} & 71.0 {(+0.9)} & 44.5 {(+2.4)} & 31.1 {(+0.6)}\\
             & & & + \textbf{FreqAug-T}  & 90.4 {(+1.4)} & 62.2 {(+2.7)} & 68.8 {(-1.3)} & 44.3 {(+2.2)} & 31.5 {(+1.0)}\\
            \hline
        \end{tabular}}
\end{center}
\caption{\textbf{Backbone comparison on Mini-Kinetics.} We extensively evaluate FreqAug with diverse backbones, including SlowOnly-50 (SO-50), SlowOnly-18 (SO-18), 3D-ResNet-18 (R-18), R(2+1)D and S3D-G, as the baselines for finetuning and low-shot learning tasks. Here, $T$: number of frames, $\tau$: input stride, $H\times W$: spatial resolution. Input resolutions are for the downstream task. }
\label{table:ax_backbone}
\end{table*}

\subsection{Additional Backbones}
\label{subsec:ax_backbone}
In Table~\ref{table:ax_backbone}, we present extended results of different backbones from Table~\ref{table:backbone}: SlowOnly-50 (SO-50), SlowOnly-18 (SO-18), 3D-ResNet-18 (R-18), R(2+1)D, and S3D-G, which have various input resolutions (number of frames $T$, input stride $\tau$, spatial resolution $H\times W$), depth, and network architecture.
The detailed setup for each backbone is described as follows: 
1) SO-50 with a low spatial resolution of $128\times 128$ trained with a larger batch size of 256, 
2) SO-18 chosen for testing a shallow model with plain residual blocks (not bottleneck blocks) and two different temporal resolutions ($8\times 8$, $16\times 4$), 3) R-18 which has residual blocks with full 3D convolutions and its input clip has 16 frames with stride 2, 4) R(2+1)D which has factorized 3D residual blocks, and
5) S3D-G which has inception-style 3D blocks. 
Note that we pretrain R(2+1)D and S3D-G with the input temporal resolution $16\times 4$ and then finetune with $32\times 2$.
The results in Table~\ref{table:ax_backbone} show FreqAug boosts performance in all cases regardless of spatial and temporal input resolutions and the network architecture except finetuning S3D-G+FreqAug-T on Diving48.
However, in the other four datasets, S3D-G with FreqAug also surpasses the baseline model.
Training hyperparameters for a particular model may need to be adjusted for such a specific downstream dataset. We conjecture that the architectural elements of S3D-G, including Inception and feature gating modules, make the model less biased to the temporal low-frequency component (LFC). Also, S3D-G may require weaker augmentation as it has a much less number of parameters than SO-50.

\subsection{Additional Evaluations}
\label{subsec:ax_eval}
In Table~\ref{table:linear}, we report three additional evaluation results of MK200-pretrained SlowOnly-50 (SO-50) in the MoCo framework: 1) linear evaluation on the pretrained dataset (MK200); 2) finetuning on Gym99; 3) finetuning on SSv2.
Note that all the pretrained models are the same models presented in the main paper.

\begin{table}[!t]
\tabcolsep=0.17cm
\small

\begin{center}
\resizebox{0.8\linewidth}{!}{
		\begin{tabular}{l|c||c|c|c}
            \hline
            \multirow{2}{*}{\makecell{Method }} & \multicolumn{1}{c||}{FreqAug} & \multicolumn{1}{c|}{Linear} &
            \multicolumn{2}{c}{Finetuning}\\
            \cline{3-5}
             & prob. & MK200 & Gym99 & SSv2 \\
            \hline
            Baseline & - & 65.3 & 88.2 & 57.2 \\
            \hline
            FreqAug-T & 0.1 &66.6 & 89.1 & 58.0 \\
            FreqAug-T & 0.3 & 67.2 & 89.5 & \textbf{58.4} \\
            FreqAug-T & 0.5 & \textbf{67.4} & 89.3 & 58.2 \\
            \hline
            FreqAug-ST & 0.1 & \textbf{67.4} & 88.8 & 57.8 \\
            FreqAug-ST & 0.3 & 66.9 & 89.1 & 58.1 \\
            FreqAug-ST & 0.5 & 67.3 & \textbf{89.6} & \textbf{58.4} \\
            \hline
        \end{tabular}}
\end{center}
\caption{\textbf{Additional evaluations of MoCo with FreqAug.} Results of linear evaluation on the pretrained dataset (MK200) and finetuning (Not low-shot) on relatively large datasets (Gym99 and SSv2) are presented. Three augmentation probabilities for FreqAug-T and FreqAug-ST are evaluated. All backbones are SO-50.}
\label{table:linear}
\end{table}

{\noindent \bf Linear Evaluation.}
Linear evaluation protocol, which trains a linear classifier on top of a frozen encoder, is often applied to the pretrained dataset for evaluating the learned representation by video SSL~\cite{Feichtenhofer2021largescale}.
The linear classifier is trained for 60 epochs with the consine learning rate schedule, zero weight decay, and the linear warm-up.
The base learning rate for the schedule is set to 0.5.
The warm-up starts from 0.001 and lasts for the first 8 epochs.
The data augmentations during training and evaluation protocols are the same as the finetuning setting on UCF101 and HMDB51.
Similar to other downstream tasks, MoCo with FreqAug surpasses the baseline for all the augmentation probabilities ($p$), which implies the learned representations get more discriminative via FreqAug.

{\noindent \bf Finetuning on Gym99 and SSv2.}
The finetuning setting used for Gym99 is identical to other datasets in the main paper.
Since SSv2 is a relatively large dataset, we follow the training recipe in the previous work~\cite{Feichtenhofer2021largescale}.
The models are finetuned for 22 epochs with the initial learning rate of 0.06, weight decay of 1e-6, and batch size of 64.
We use the step-wise learning rate decay at 14 and 18 epoch by $1/10$.
We set the dropout probability to 0.5.
We observe again that FreqAug improves the performance over the baseline regardless of the augmentation probability.
The trend that FreqAug with higher $p$ usually shows better accuracy is similar to the low-shot learning in the main paper.
However, the difference between them is less clear than the low-shot learning results, presumably because of the relatively large dataset size of the downstream task, as we claimed in the main paper. 

\begin{figure}[!t]
\centering
    \includegraphics[width=\linewidth]{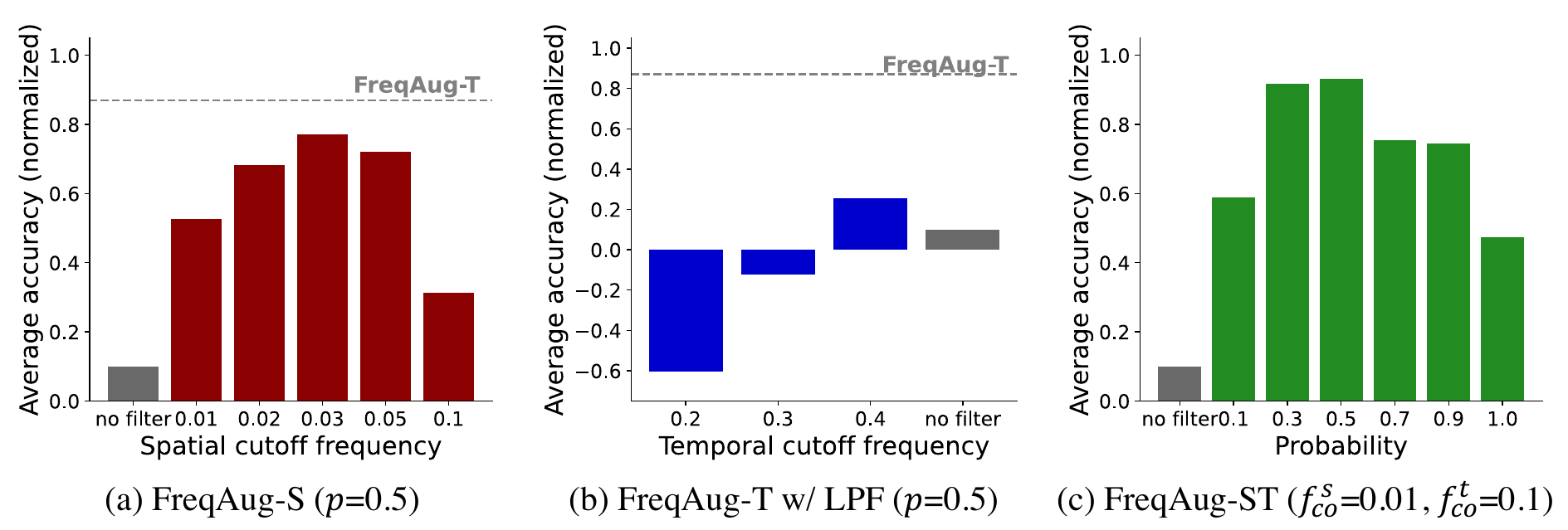}
    \caption{\textbf{Additional FreqAug hyperparameter ablations on Mini-Kinetics.} Three parameters are searched: (a) spatial cutoff frequency for FreqAug-S and (b) temporal cutoff frequency for FreqAug-T with low-pass filter (LPF) and (c) augmentation probability for FreqAug-ST. Other parameters is set fixed as in the parenthesis. Min-max normalized accuracies of 5 tasks are averaged; min/max values calculated for Fig.~\ref{fig:ablation} are used for better comparison.}
\label{fig:ax_ablation}
\end{figure}

\begin{figure}[!t]
\centering
    \includegraphics[width=\linewidth]{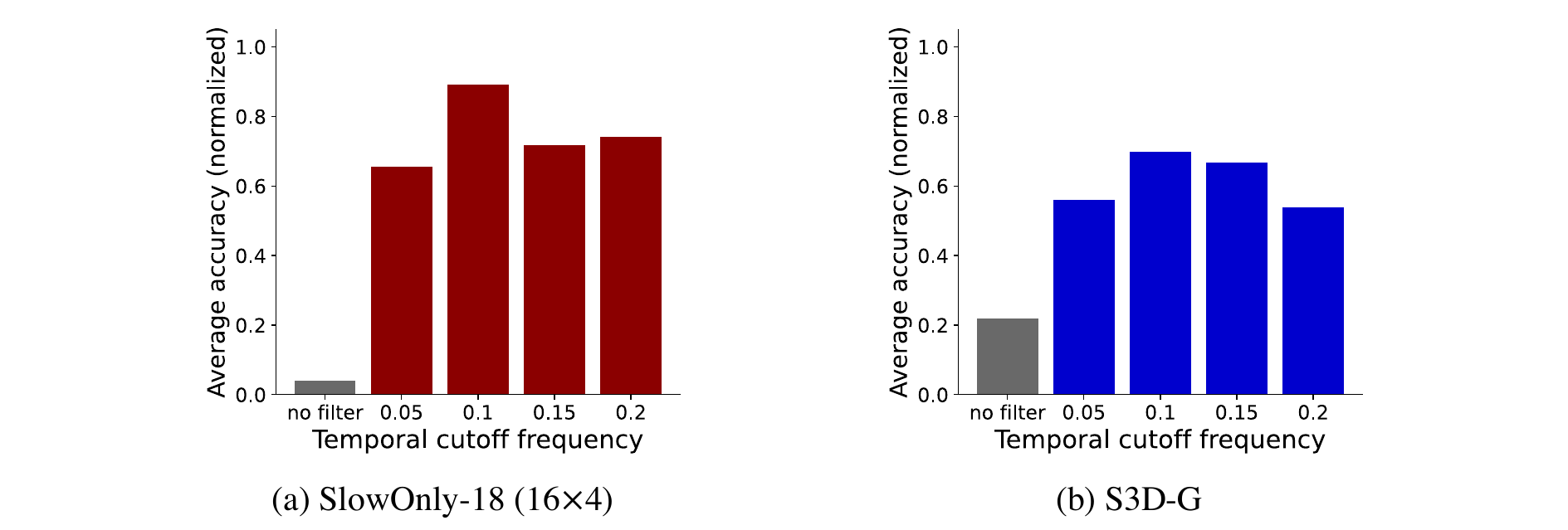}
    \caption{\textbf{Additional ablation study for backbones with more input frames}: (a) SlowOnly-18 (16$\times$4) and (b) S3D-G. Different temporal cutoff frequencies are tested for FreqAug-T on Mini-Kinetics. Augmentation probability is set fixed to 0.5. Min-max normalized accuracies of 5 tasks are averaged; min/max values calculated for each backbone.}
\label{fig:ax_ablation_16x4}
\end{figure}
\subsection{Ablation Studies on Hyperparameters}
\label{subsec:ax_ablation}
{\noindent \bf Description of Figure~\ref{fig:ablation}.}
Here, we explain in detail about Fig.~\ref{fig:ablation} in the main paper.
First, we search four cutoff frequencies $f^t_{co}{\in}\{0.1,0.2,0.3,0.4\}$ for temporal HPF with augmentation probability $p{=}0.5$ in Fig.~\ref{fig:ablation} (a).
The first three cases show improved downstream performance compared to the baseline (``no filter" in the figure), with notable improvement.
Because $f^t_{co}{=}0.1$ is more dominant than others on average, we set it as a default.
If $f^t_{co}$ becomes larger, then the performance increment over the baseline becomes smaller since too much information is removed.
Second, $p$ of FreqAug-T is searched in a range of $[0.0, 1.0]$ while keeping $f^t_{co}$ fixed to 0.1 in Fig.~\ref{fig:ablation} (b).
We found that the overall downstream performance is the best when $p{=}0.5$, which is the default setting of FreqAug-T.
In addition, FreqAug-T enhances the performance regardless of $p$; even in the case $p{=}1.0$ where temporal HPF is always applied. 
Lastly, we examine FreqAug-ST by adding spatial HPF on top of FreqAug-T with default hyperparameters found above. 
The spatial cutoff frequencies $f^s_{co}{\in}\{0.01,0.03,0.05\}$ are tested in Fig.~\ref{fig:ablation} (c).
We observed that FreqAug-ST with $f^s_{co}{=}0.01$ improves the performance even more than FreqAug-T, while a larger $f^s_{co}$ has a negative impact on the performance.  
More ablation studies can be found in the following.

{\noindent \bf Additional Ablation Studies.}
In addition to Fig.~\ref{fig:ablation}, we conduct additional ablation studies on FreqAug hyperparameters in Fig~\ref{fig:ax_ablation}.
Note that we use the same min/max values of Fig.~\ref{fig:ablation} in the main paper for the normalization.
Therefore, some scores can be below zero if one model's accuracy is lower than the lowest accuracy of the models in Fig.~\ref{fig:ablation}.

First, we examine spatial-filter-only variants of FreqAug (FreqAug-S) in Fig.~\ref{fig:ax_ablation} (a) by testing several $f^s_{co}$.
We observe that FreqAug-S also outperforms the baseline, but it is worse than the default FreqAug-T.
Second, accuracies of the low-pass filter (LPF) counterpart of FreqAug-T with different $f^t_{co}$ are displayed in~\ref{fig:ax_ablation} (b).
The accuracy becomes even less than the baseline as $f^t_{co}$ gets smaller, \ie, as more high-frequency components are removed.
The results clearly demonstrate that the performance improvement via FreqAug with HPF is not just due to andomness; each frequency component has a different influence on the downstream performance.
Lastly, $p$ of FreqAug-ST is searched in a range of $[0.0, 1.0]$ while keeping $f^s_{co}$ and $f^t_{co}$ fixed in Fig.~\ref{fig:ax_ablation} (c).
We found that the trend is similar to the case of FreqAug-T in Fig. 3(b) of the main paper; the overall downstream performance is the best when $p{=}0.5$.

In Fig.~\ref{fig:ax_ablation_16x4}, we also investigate the effect of changing $f^t_{co}$ for FreqAug-T on backbones with a larger number of input frames: (a) SO-18 and (b) S3D-G. These models take input videos with 16 frames during pretraining.
Here, we normalize the accuracy with the min/max values of models among the same backbone.
We test $f^t_{co}{\in}\{0.05,0.1,0.15,0.2\}$ and observe that models with FreqAug-T outperform its baseline on average in all test cases.
The result also shows that the smallest possible $f^t_{co}$, 0.05 in this example, is not always the best choice.
Therefore, searching the best performing $f^t_{co}$ is required depending on each backbone or input resolution.

\subsection{Ablation Studies on Views}
\label{subsec:ax_view}
\begin{table}[!t]
\tabcolsep=0.05cm
\begin{center}
\resizebox{0.9\linewidth}{!}{
		\begin{tabular}{c c |c|c|c|c|c|c}
            \hline
            \multicolumn{3}{c|}{FreqAug} & \multicolumn{3}{c|}{Finetune} &\multicolumn{2}{c}{Low-shot (10\%)}\\
            \cline{1-8}
            \multicolumn{1}{c|}{\small{view 1}} &\multicolumn{1}{c|}{\small{view 2}} & \small{$p$} & \small{UCF101} & \small{HMDB51} & \small{Diving48} & \small{Gym99} & \small{SSv2} \\
            \hline
            \multicolumn{2}{c|}{} & & 87.0 & 56.5 & 67.8 & 29.9 & 25.3\\
            \hline
            \checkmark & & 0.5 & 88.8 & 56.7 & \textbf{71.4} & 30.4 & 25.7 \\
            \checkmark & & 1.0 & 86.8 & 54.6 & 68.3 & 30.2 & 24.6 \\
            &\checkmark & 0.5 & 88.9 & 59.9 & 66.5 & \textbf{35.7} & 27.4 \\
            &\checkmark & 1.0 & 86.9 & 57.8 & 65.9 & 33.4 & 26.2 \\
            \hline
            \checkmark & \checkmark & 0.5 & \textbf{89.8} & \textbf{60.8} & 70.3 & 35.2 & \textbf{28.1} \\
            \hline
        \end{tabular}}
\end{center}

\caption{\textbf{Impact of FreqAug only on a single view.} FreqAug column indicates the views where FreqAug is applied, and the augmentation probability for FreqAug. View-2 is input for the momentum encoder. Tested with MoCo and FreqAug-T on MK200. All backbones are SlowOnly-50.}
\label{table:view}
\end{table}

In the main paper, we applied FreqAug to both views of the video SSL models as the default setting. On the other hand, we individually apply FreqAug-T to a single view of MoCo in Table~\ref{table:view}.
View-1 and view-2 indicate the inputs for the trainable encoder and the momentum encoder, respectively.
We observe that our default setting is generally better on the five action recognition downstream tasks than the single view counterparts.
Compared to the default setting, FreqAug-T on view-1 shows higher top-1 accuracy on Diving48 but significantly worse performance on the others.
FreqAug-T on view-2 shows a marginal improvement on Gym99 dataset but a considerable degradation on Diving 48.
FreqAug-T on both views achieves decent performance without a significant loss on any task.
We also provide results of single-view FreqAug with $p=1.0$ where one of the views is always filtered.
In this case, the invariance between two views with all frequency components cannot be learned.
The reduced performances of the case $p=1.0$ support our claim that invariance learning between different frequency components needs to be adjusted appropriately by $p$, rather than learning from \textit{only} low- or high-frequency components.
Further analysis on the impact of applying FreqAug to each stream on each downstream task remains for future research.

\subsection{Localization Evaluation of BYOL Features}
\label{subsec:ax_byol}
\begin{table}[!t]
\tabcolsep=0.1cm

\begin{center}
\resizebox{0.95\linewidth}{!}{
		\begin{tabular}{l|c|c|c|c|c|c}
            \hline
            Method & Data & Acc. & Edit & \multicolumn{3}{c}{F1@\{0.10, 0.25, 0.50\}}\\
            \hline
            BYOL & \multirow{3}{*}{\makecell{K400}} & \hspace*{1.5mm}59.5\hspace*{1.5mm} & \hspace*{1.5mm}63.0\hspace*{1.5mm} & \hspace*{1.5mm}59.4\hspace*{1.5mm} & \hspace*{1.5mm}53.9\hspace*{1.5mm} & \hspace*{1.5mm}42.0\hspace*{1.5mm}\\
            \ \ + \textbf{FreqAug-ST} &  & 61.9 & 64.3 & 62.0 & 56.8 & 44.9 \\
             \ \ + \textbf{FreqAug-T} & & \textbf{64.6} & \textbf{65.1} & \textbf{62.1} & \textbf{56.9} & \textbf{45.4}  \\
            \hline
        \end{tabular}}
\end{center}
\caption{
\textbf{Temporal action segmentation with BYOL pretrained features on Breakfast.} 
All features are evaluated with MS-TCN. 
`Edit' denotes edit distance.
Scores are averaged over 10 evaluations on split-1.
}
\label{table:ax_seg}
\end{table}

\begin{table}[!t]
\tabcolsep=0.1cm

\begin{center}
\resizebox{1.0\linewidth}{!}{
		\begin{tabular}{l|c|c|c|c|c|c|c}
            \hline
            Method & Pretrain & \multicolumn{5}{c|}{mAP@\{0.3, 0.4, 0.5, 0.6, 0.7\}} & Avg\\
            \hline
            BYOL & \multirow{3}{*}{\makecell{Self-sup.}} & \hspace*{1.5mm}53.3\hspace*{1.5mm} & \hspace*{1.5mm}46.7\hspace*{1.5mm} & \hspace*{1.5mm}37.6\hspace*{1.5mm} & \hspace*{1.5mm}27.7\hspace*{1.5mm} & \hspace*{1.5mm}17.9\hspace*{1.5mm} & 36.6 \\
             \ \ + \textbf{FreqAug-ST} &  & 53.9 & 46.6 & 38.2 & 27.9 & 17.8 & 36.9 \\
             \ \ + \textbf{FreqAug-T} &  & \textbf{54.4} & \textbf{47.9} & \textbf{39.1} & \textbf{28.9} & \textbf{18.8} & \textbf{37.8} \\
            \hline
        \end{tabular}}
\end{center}
\caption{
\textbf{Temporal action localization with BYOL pretrained features on THUMOS'14.} 
Features are pretrained on K400 and evaluated with G-TAD. 
Scores are mean over 5 runs.
}
\label{table:ax_loc}
\end{table}

In Table~\ref{table:ax_seg} and~\ref{table:ax_loc}, we evaluate pretrained BYOL features on temporal action segmentation and localization downstream tasks as in Table~\ref{table:seg} and Table~\ref{table:loc}, respectively.
Similar to the case of MoCo, we observe that FreqAug-enhanced BYOL features outperform features of the baseline BYOL in all metrics.
Interestingly, the downstream performance of BYOL-pretrained features are similar to that of MoCo-pretrained one, unlike the case in action recognition downstream tasks where BYOL outperforms MoCo.
This is why we need diverse downstream tasks for evaluating various pre-training methods.
In image domain, a similar observation was reported in~\cite{Ericsson2021howwell}.

\section{Additional Discussions}
\label{sec:ax_discuss}
\subsection{t-SNE Analysis on Pre-trained Features}
\label{subsec:ax_tsne}
\begin{figure}[!t]
  \centering
   \includegraphics[width=\linewidth]{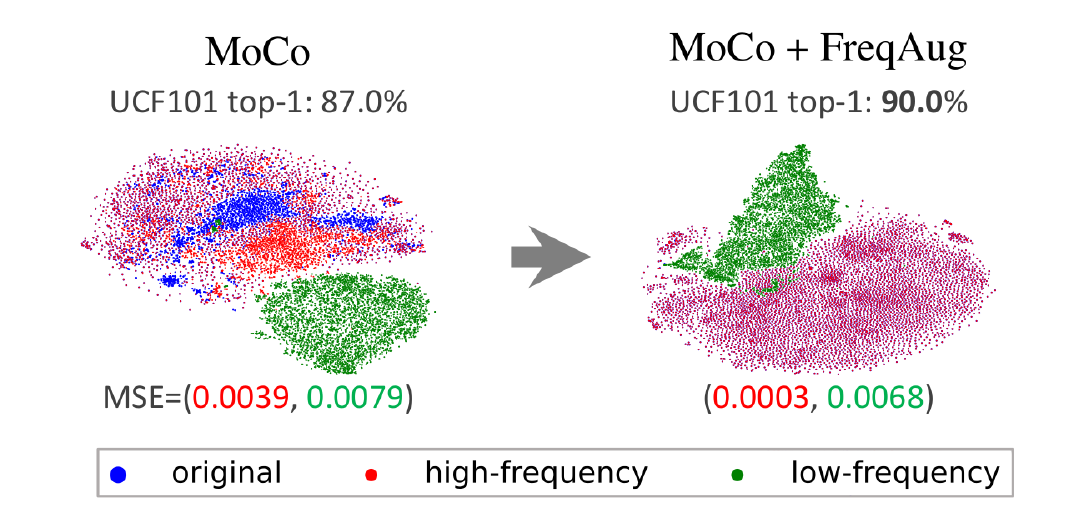}
   \caption{\textbf{t-SNE visualization of the output features from original frames (blue) and its spatial HFC (red) or LFC (green).} Mean squared error (MSE) between original features with HFC/LFC are presented under each plot. MoCo pretrained SlowOnly-50 models with or without FreqAug (and UCF101 finetuning acccuracies) are compared. FreqAug makes features of HFC close to that of original clips which results in better downstream performance. If red and blue dots are too close, they can be perceived as purple.}
   \label{fig:ax_tsne_s}
\end{figure}
{\noindent \bf Spatial Dimension.}
In addition to Fig.~\ref{fig:tsne} in the main paper, we also visualize t-SNE plots of features of original clips (blue) and the clip with either spatial HFC/LFC (red/green) in Fig.~\ref{fig:ax_tsne_s}.
We set $f^s_{co}=0.02$ for both HPF and LPF.
Similar to the temporal dimension case, MSE between original clip features and its spatial HFC is reduced considerably because of FreqAug.
It also implies that the model pretrained with FreqAug learns more invariant features on spatial LFC than the baseline.

\begin{figure}[!t]
\centering
    \includegraphics[width=0.95\linewidth]{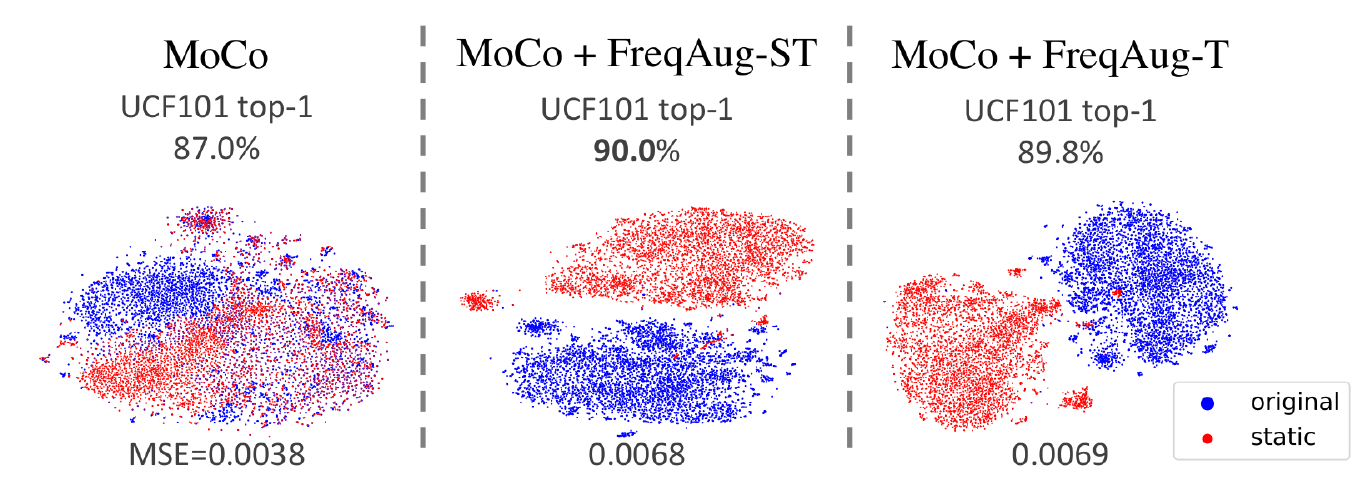}
    \caption{\textbf{t-SNE visualization of the output features from the original (blue) and static (red) videos.} A static video is a synthesized video with repetitions of the center frame without any temporal change. Mean squared error (MSE) between two features is presented under each plot. MoCo pretrained SlowOnly-50 models with or without FreqAug (and their UCF101 finetuning accuracies) are compared. FreqAug makes features of the static videos far from that of original videos which results in better downstream performance. When red and blue dots are too close, they can be perceived as purple.}
\label{fig:ax_tsne}
\end{figure}
{\noindent \bf Static Videos.}
In Fig.~\ref{fig:ax_tsne}, we also visualize the distance between the model's output feature of the original videos and that of the static videos.
Here, a static video indicates a synthesized video with duplicated center frames; thus, there is no temporal information; it may or may not have important spatial features depending on the frame.
We compare three features from MoCo-pretrained SlowOnly-50 on MK200: the baseline, with FreqAug-ST, and with FreqAug-T.
We use mean squared error (MSE) as the distance metric, and use videos in the validation set of MK200 for the visualization.
We observed that the two feature distributions are much closer for the MoCo baseline than for the models trained with FreqAug.
It implies that the baseline model extracts more spatial features from the original video since static videos have only spatial cues.
On the other hand, FreqAug makes the two distributions apart.
It means the models trained with FreqAug extract fewer features from static videos, which have non-zero components only where the temporal frequency is zero.
We believe that the feature representation of the model trained with FreqAug whose temporal zero-frequency components have been weakened, can affect the downstream performance of the model.

\subsection{Description of Figure~\ref{fig:gradcam} (a)}
\label{subsec:ax_group}
We calculate the amplitude of each frequency component in the spectrum and get the relative amount of low-frequency components (LFC), which is the amplitude of target frequency components divided by the sum of the amplitude of the entire spectrum. We sort the samples in increasing order according to the proportion of LFC, then split the dataset into several groups in order with an equal number (210-211) of samples. We choose the temporal and spatial zero-frequency components as the LFC. We plot the accuracy difference between the MoCo pretrained model with and without FreqAug for each bin; accuracy of MoCo+FreqAug minus accuracy of MoCo baseline. Bins with a larger low-frequency ratio are located on the right on the horizontal axis of the plot. 

\begin{figure*}[!t]
\centering
    \includegraphics[width=0.95\linewidth]{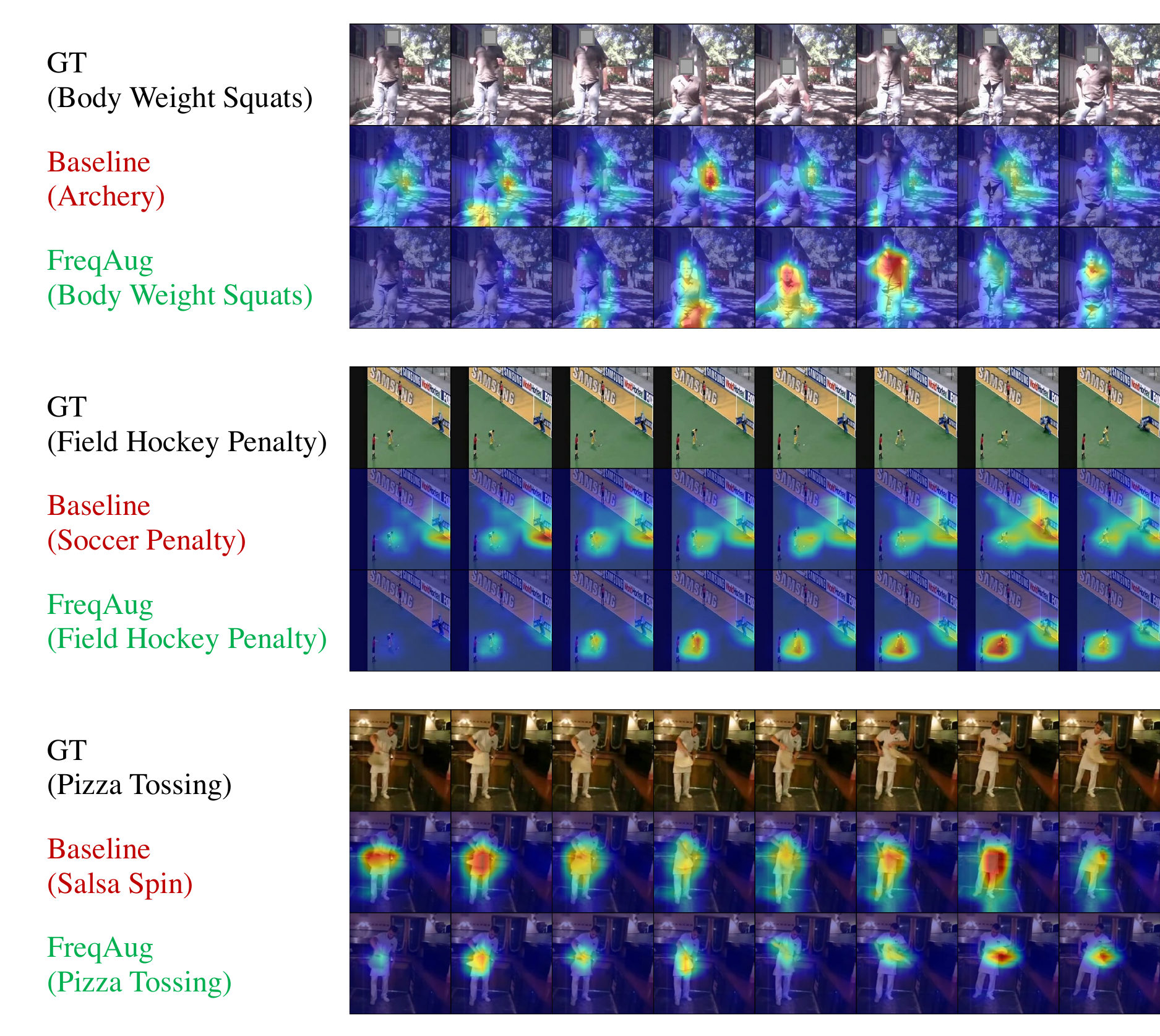}
    \caption{
    \textbf{GradCAM on UCF101.} 
    Finetuning models from MoCo (Baseline) and MoCo with FreqAug (FreqAug) pretraining are compared. The ground truth (GT) labels and the model's predictions are presented in the parenthesis.}
\label{fig:ax_gradcam}
\end{figure*}
\subsection{More GradCAM Visualization}
\label{subsec:ax_gradcam}
In Fig.~\ref{fig:ax_gradcam}, we visualize three additional GradCAM examples from the bins with a large low-frequency ratio.
These selected samples show what information the model is focusing on.
In the first sample, FreqAug model attends to the man squatting while Baseline model focuses on the background.
Similarly, in the second sample, activation of Baseline is on the field and the goal post, which can be confused with soccer, while that of FreqAug is on the players which are very small part of entire frames.
In the last video, FreqAug model could successfully recognize the action by focusing on the small pizza dough and the upper body of the person rather than on the entire body.
To take the analysis and visualization (including Sec.~\ref{subsec:tsne}) into account, we can conclude that FreqAug helps the model to focus on active or motion-related areas in the videos with static backgrounds, which likely have a high proportion of temporal low-frequency components. 

\begin{figure}[!t]
\centering
    \includegraphics[width=\linewidth]{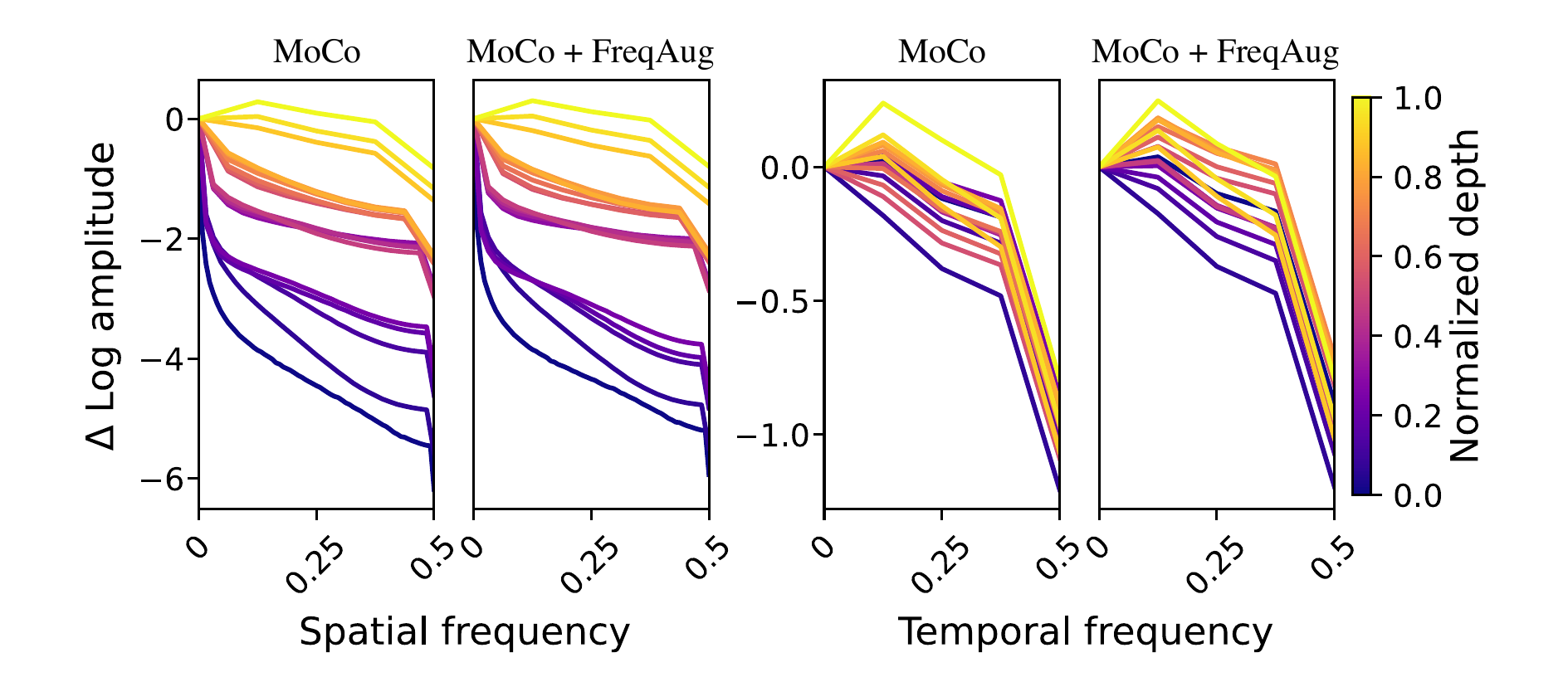}
    \caption{\textbf{Relative log amplitude of frequency components in feature maps: spatial (left) and temporal (right) axis.} 
    Color represents the normalized depth of each layer (lower value means closer to the input). 
    $\Delta$ Log amplitude is the difference between the log amplitude at each frequency and at zero-frequency (0.0). 
    Features from MoCo-pretrained SlowOnly-50 encoders are visualized. 
    MoCo with FreqAug-T increases the relative amplitude, especially for temporal frequency components, in some intermediate layers.}
\label{fig:amp}
\end{figure}
\subsection{Analysis on Frequency Amplitude in the Feature Map}
\label{subsec:ax_amp}
For analyzing the effect of FreqAug on the feature, we choose a method that visualizes the relative log amplitude of frequency components of the feature map in Fig.~\ref{fig:amp}. 
The method is originally proposed to compare convolution and self-attention operations~\cite{park2022how}.
The feature map for each layer is displayed in different colors according to the depth of the layer.
The depth is normalized for visualization, so 0.0 indicates the first layer, and 1.0 indicates the last layer.
The vertical axis of the graph represents the log amplitude at each frequency relative to the log amplitude at the zero frequency by substituting it.
The horizontal axis denotes normalized frequency along the spatial or temporal axis.
Two models, MoCo pretrained SO-50 with and without FreqAug-T, are compared.
First, we found that relative amplitude for MoCo baseline model shows different trends in spatial and temporal frequency.
In the spatial axis, relative frequency amplitudes get larger as the signal goes through the stages (a set of layers), while the amplitudes are rather mixed in the temporal axis.
Second, when FreqAug-T is added, high-frequency (larger than zero) amplitudes of some intermediate layers increase especially in the temporal axis. 
We believe that FreqAug-T affects the temporal frequency of the middle layers because only later stages of SO-50 contain temporal convolutions for the temporal modeling.
This analysis supports our claim that FreqAug makes the model exploit relatively more high-frequency components by showing a relative increase of high-frequency amplitudes in intermediate feature maps.

\begin{table}[!t]
\tabcolsep=0.13cm

\begin{center}
\resizebox{0.95\linewidth}{!}{
		\begin{tabular}{c|c|c|c|c|c|c}
            \hline
             & \multirow{2}{*}{\makecell{Method}} & \multicolumn{3}{c|}{Finetune} &\multicolumn{2}{c}{Low-shot (10\%)}\\
            \cline{3-7}
            & & UCF101 & HMDB51 & Diving48 & Gym99 & SSv2 \\
            \hline
             & Baseline & 87.0 & 56.5 & 67.8 & 29.9 & 25.3\\
             & \textbf{FreqAug-ST} & \textbf{90.0} & \textbf{61.6} & \textbf{71.0} & \textbf{34.8} & 28.1\\
            \hline
            \multirow{6}{*}{(a)} & GF (k=3, $\sigma$=0.5) & 88.7 & 56.9 & 69.7 & 33.4 & 27.0\\
            & GF (k=3, $\sigma$=2.0) & 88.3 & 58.4 & 68.9 & 33.4 & 26.0\\
            & GF (k=7, $\sigma$=0.5) & 88.9 & 58.7 & 68.3 & 31.7 & 27.4\\
            & GF2D (k=3, $\sigma$=0.5) & 87.0 & 58.2 & 67.1 & 31.5 & \textbf{28.9}\\
            & GF2D (k=3, $\sigma$=2.0) & 86.9 & 58.6 & 66.3 & 31.9 & 27.4\\
            & GF2D (k=7, $\sigma$=0.5) & 87.6 & 59.2 & 68.4 & 32.4 & 28.1\\
            \hline
            \multirow{4}{*}{(b)} & RandConv & 88.4 & 61.5 & 67.0 & 34.5 & 28.4\\
            & RandConv3D & 87.2 & 57.8 & 69.0 & 32.3 & 26.8\\
            & Mixup (InputMix) & 85.8 & 54.4 & 69.0 & 31.9 & 24.2\\
            & Frame Mixup (BE) & 88.8 & 57.5 & 68.2 & 30.8 & 25.4\\
            & AM($\eta$=0.2) & 87.4 & 56.5 & 68.9 & 31.0 & 25.5\\
            & AM($\eta$=1.0) & 85.4 & 52.8 & 68.5 & 31.4 & 24.6\\
            \hline
        \end{tabular}}
\end{center}
\caption{\textbf{Comparison with other methods:} (a) Gaussian filters (GF) with different kernel sizes (k) and std. ($\sigma$) in spatio-temporal domain and (b) other augmentations. All methods are tested with MoCo-pretrained SlowOnly-50 on MK200.}
\label{table:ax_compare}
\end{table}

\subsection{Comparison with Spatio-temporal Filter}
\label{subsec:ax_stfilter}
One may wonder the reason for filtering in the frequency domain rather than in the spatio-temporal domain.
For example, Gaussian blurring, \ie, convolution of Gaussian kernel, is a simple low-pass filter (LPF) in the spatio-temporal domain.  
First, designing filters in the frequency domain is more intuitive and definite, especially for multi-dimensional signals, since the desired frequency band can be chosen simply by multiplying a filter mask.
We also empirically test 3D Gaussian HPF (GF) with a few kernel sizes (k) and standard deviations ($\sigma$) in Table~\ref{table:ax_compare} (a).
GF2D is a variant of GF with a spatial-only kernel.
Both GF and GF2D mostly show improvements over the baseline but not as much as FreqAug.
Regarding training time, MoCo+FreqAug is around 10\% faster than MoCo+GF($k$=3) on 4 NVIDIA P40 GPUs (880, 981 sec/epoch on MK200, respectively).
These imply that filtering in the frequency domain may also have advantages in terms of computational performance depending on the type of the spatio-temporal filter.

\subsection{Comparison with Other Augmentation Methods}
\label{subsec:ax_other}
We compare FreqAug with several other regularization methods: RandConv (RC\textsubscript{img1-7,p=0.5} in~\cite{xu2021robust}), Mixup (InputMix in~\cite{lee2021imix}), static frame Mixup (BE~\cite{wang2021removing}), and Amplitude Mix (AM~\cite{Xu2021FACT}) in Table~\ref{table:ax_compare} (b).
According to Fig. 1 in the original paper~\cite{xu2021robust}, it seems that RandConv modulates the input signal while preserving some high-frequency components. RandConv3D is a direct extension of 2D kernels into 3D kernels with the same hyperparameters.
We choose $\eta$ of 0.2 and 1.0 for AM used in the paper.
Note that we only apply the augmentation on the input video and do not apply any changes in the model or the loss for the fair comparison.
RandConv, BE, and AM mostly improve the performance over the baseline while InputMix affects adversely.
Previous work (Tab. 4 in~\cite{wang2021removing}) also shows that naive Mixup is not helpful for downstream performance.
Overall results show the superiority of our methods over other augmentations in video SSL frameworks.

\subsection{Comparison with Optical Flow}
\label{subsec:ax_of}
Both optical flow and our method extract temporally changing patterns from the visual signal.
Though optical flow is a powerful motion feature, using it in SSL is somewhat cumbersome.
Aside from the cost of extracting the optical flow from the video, a dedicated encoder is usually required for processing it.
If one wants to utilize optical flow as an additional branch, it will result in increased computation.
Or, to replace a few RGB branches with optical flow branches is not always straightforward in some SSL methods.
On the other hand, FreqAug can be seamlessly integrated into any RGB-based SSL method.

\begin{figure*}[!t]
\centering
    \includegraphics[width=0.95\linewidth]{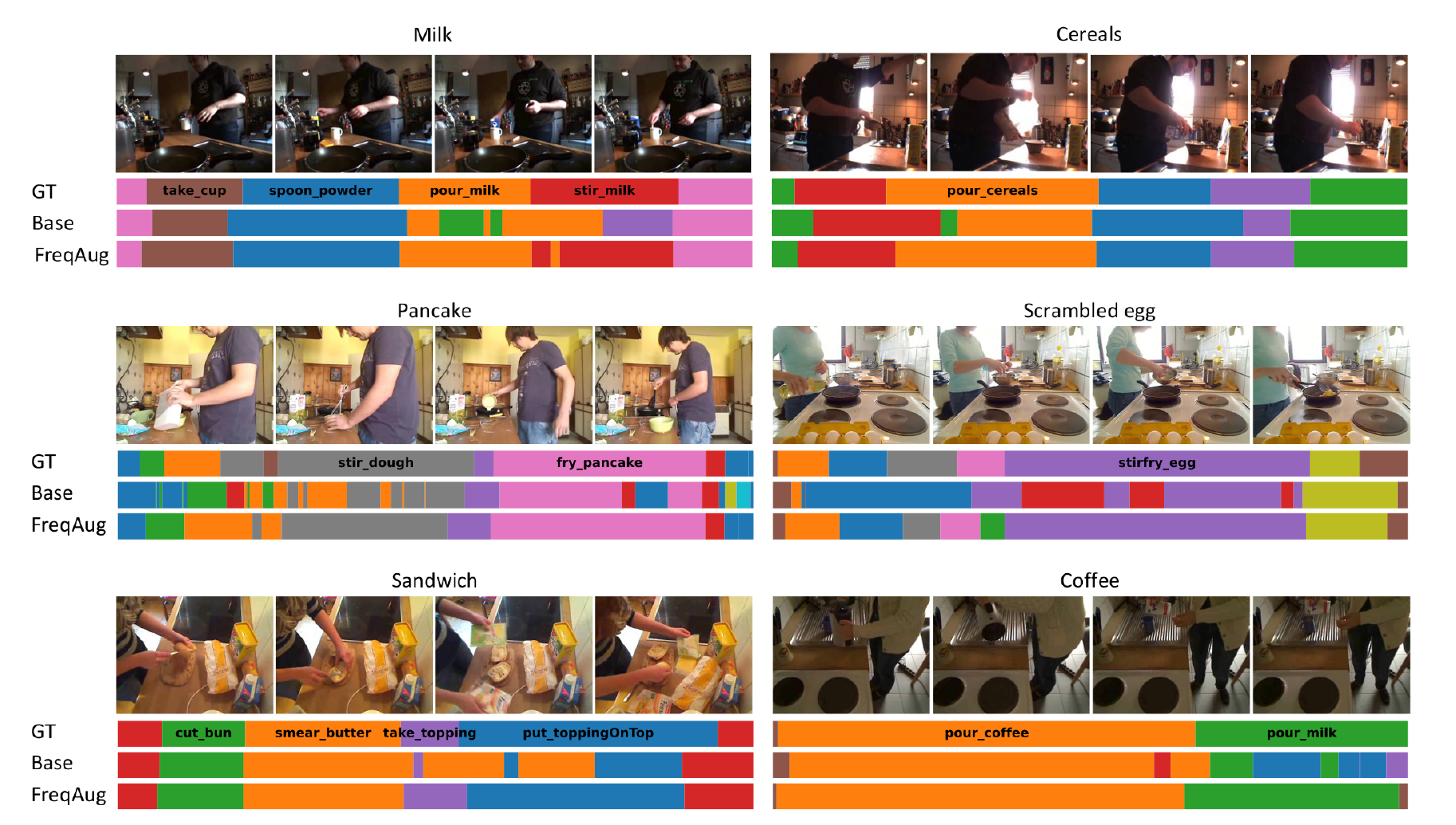}
    \caption{
    \textbf{Qualitative results of temporal action segmentation on the Breakfast dataset.} 
    Bar plot displays predictions of pretrained features with the baseline augmentation (Base) and FreqAug-T (FreqAug) along with the ground-truth (GT). SlowOnly-50 encoders are pretrained by MoCo framework and MS-TCN is trained on top of frozen features from the pretrained encoder.}
\label{fig:seg}
\end{figure*}
\subsection{Qualitative Analysis of Action Segmentation}
\label{subsec:ax_seg}
The qualitative results of the temporal action segmentation task are presented in Fig.~\ref{fig:seg}.
Two features pretrained by MoCo with the baseline augmentation (Base) and 
with FreqAug-T (FreqAug) can be compared with the ground-truth (GT) label.
The quality of the segmentation with FreqAug is much better than the baseline; the classification and the boundaries of actions are more precise.
As discussed in the main paper, we observe that the background scene in the Breakfast dataset rarely changes, so it makes the learned representation with FreqAug more effective on this task.

\begin{figure*}[!t]
\centering
    \includegraphics[width=\linewidth]{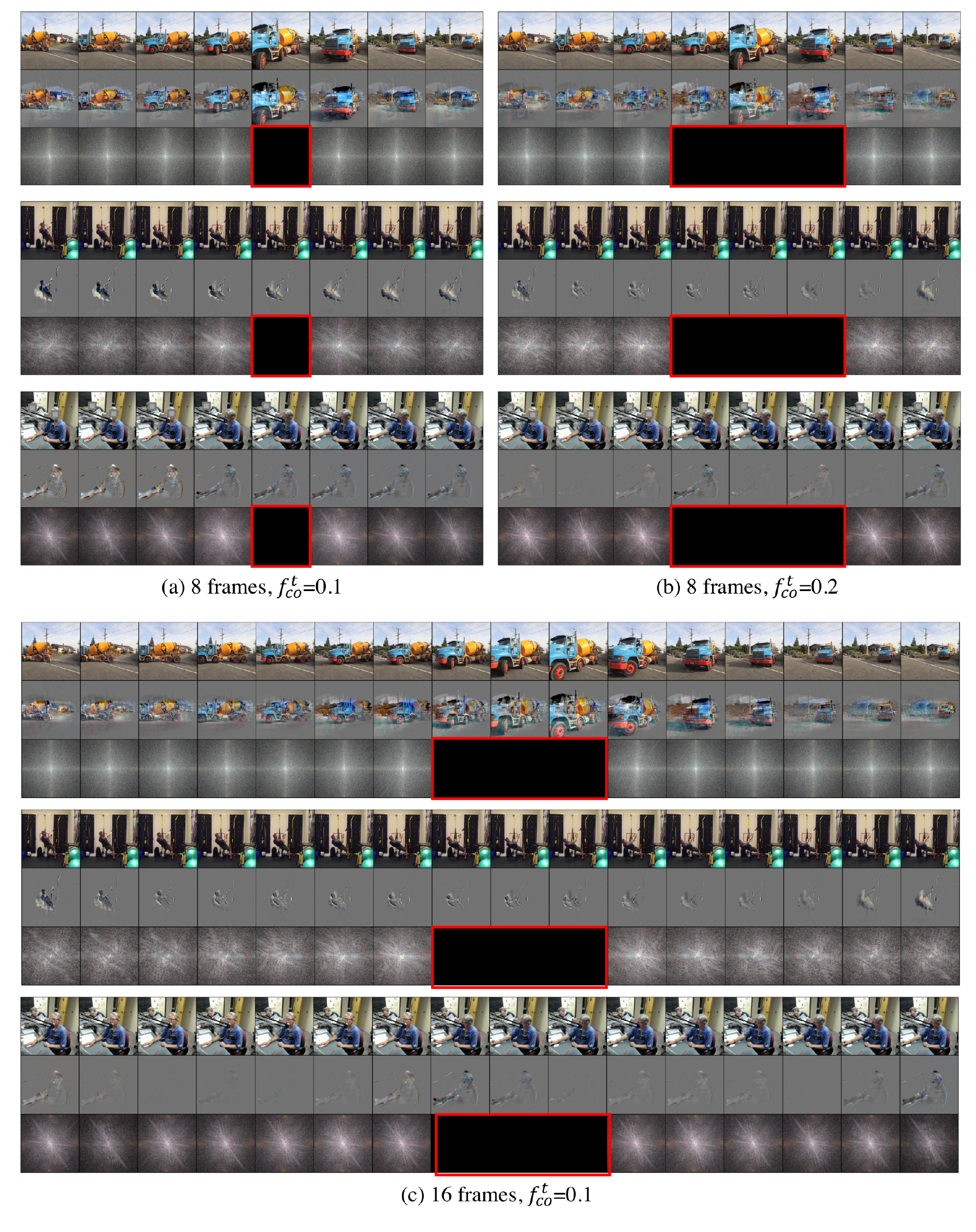}
    \caption{
    \textbf{More examples of filtered frames and spectrum.}
    Three different settings (number of frames (T) and temporal cutoff frequency ($f^t_{co}$)) for temporal high-pass filter (HPF) are displayed: (a) $T=8$, $f^t_{co}=0.1$ (default), (b) $T=8$, $f^t_{co}=0.2$, and (c) $T=16$, $f^t_{co}=0.1$. 
    Top, middle, and bottom rows of each sample denote original frames, filtered frames, and filtered spectrum, respectively. 
    In the spectrum, the red box indicates where the temporal frequency is filtered.
    }
\label{fig:filter}
\end{figure*}
\section{More Visualizations of Temporal Filtering}
\label{subsec:ax_filter}
In Fig.~\ref{fig:filter}, we present more visualization of temporal HPF with different number of input frames ($T$), input stride ($\tau$), and temporal cutoff frequencies ($f^t_{co}$): (a) $T\times\tau{=}8\times8$, $f^t_{co}{=}0.1$ (default), (b) $T\times\tau{=}8\times8$, $f^t_{co}{=}0.2$, and (c) $T\times\tau{=}16\times4$, $f^t_{co}{=}0.1$.
Three rows of each sample display original frames, filtered frames, and filtered spectrum in order. 
In the spectrum, the red box indicates the temporal frequencies which are filtered out.
Temporal HPF with default setting (a) shows that static parts of the frames are attenuated, similar to the examples in the main paper.
When $f^t_{co}$ becomes larger (b), more portions of the spectrum are filtered out, and accordingly, more visual information is removed.
As we found in Fig. 3(a) in the main paper, removing too much visual information might cause inferior results.
In (c), more frequency components are masked, though $f^t_{co}$ is not changed, since frequencies are more densely sampled as $T$ gets larger.
This can result in a more attenuated signal in the spatio-temporal domain.

\end{document}